\documentclass[runningheads]{llncs}
\usepackage[T1]{fontenc}
\usepackage{booktabs}
\usepackage{graphicx}
\usepackage{array}
\usepackage{url}
\usepackage{tikz}
\usepackage{float}
\usepackage{longtable}
\usepackage{adjustbox}
\usepackage{xcolor}
\usepackage{todonotes}

\definecolor{addedcolor}{RGB}{0, 100, 0} 
\definecolor{deletedcolor}{RGB}{200, 0, 0} 

\usetikzlibrary{positioning}
\usetikzlibrary{arrows.meta} 
\usepackage{etoolbox}
\usepackage{comment}

\newtoggle{extended}
\toggletrue{extended} 

\iftoggle{extended}{
  \newenvironment{onlyextended}{\ignorespaces}{\ignorespacesafterend}
}{
  \excludecomment{onlyextended}
}

\iftoggle{extended}{
  \excludecomment{onlyshort}
}{
  \newenvironment{onlyshort}{\ignorespaces}{\ignorespacesafterend}
}

\newenvironment{inboth}{}{}

\newcommand{\ppinfo}[1]{{ \normalfont\color{red}\small(#1 pp)}}
\renewcommand{\ppinfo}[1]{} 

\begin{document}
\title{Knowledge Graph Re-engineering Along the Ontological Continuum\iftoggle{extended}{
\\{\normalfont(Extended version)}
}{}}
\titlerunning{KG Re-engineering Along the Ontological Continuum}
%
\author{Enrico Daga\inst{1}\orcidID{0000-0002-3184-5407} \and
Valentina Tamma\inst{2}\orcidID{0000-0002-1320-610X} \and
Terry R. Payne\inst{2}\orcidID{0000-0002-0106-8731}}
\authorrunning{E. Daga et al.}
%
\institute{The Open University, Walton Hall, Milton Keynes, United Kingdom \email{enrico.daga@open.ac.uk} \and
School of Computer Science and Informatics, University of Liverpool, UK
\email{\{V.Tamma,T.R.Payne\}@liverpool.ac.uk}}
\maketitle     


\begin{onlyshort}
\begin{abstract}
\vspace{-0.3cm}
Knowledge graphs have become the primary vehicle for data integration and are critical to modern AI. 
Yet, the diversity of KG modelling practices, from lightweight vocabularies to richly axiomatised ontologies, makes integration and reuse expensive and brittle. 
This challenge is particularly acute for GenAI-driven automation of knowledge engineering, which navigates the KG space without precise directions. 
We introduce the \emph{ontological continuum}: a task-relative navigation framework defined by two orthogonal distinctions: semantics versus pragmatics, and properties versus affordances, yielding four dimensions along which KGs can be described, compared, and transformed.
Our methodological stance is empirical: we propose the ontological continuum as a theory of the existent, derived from observation of real-world KG engineering practices, whose structure can be made formally explicit, for example, through Formal Concept Analysis (FCA). 
We ground the vision in a case study on provenance knowledge and articulate five open research challenges for the community.   
\vspace{-0.3cm}
\end{abstract}
\end{onlyshort}

\begin{onlyextended}
\textit{{\small~This paper is an extended version of a vision paper accepted at the 25th International Conference on Knowledge Engineering and Knowledge Management (EKAW), September 29th - October 1st, 2026, University of Torino, Italy.}}    
\end{onlyextended}
\begin{onlyextended} \begin{abstract}
Knowledge graphs have become the primary vehicle for data integration and are critical to the success of modern AI — but the diversity of KG modelling practices, from lightweight vocabularies to richly axiomatised ontologies, makes integration and reuse expensive and brittle. 
This challenge is particularly acute in neuro-symbolic AI, where bridging neural and symbolic components depends on the ability to re-engineer KGs to fit new requirements; GenAI now offers unprecedented automation capability, but without a principled understanding of the KG engineering space, such automation remains conceptually ungrounded. 

We introduce the \emph{ontological continuum} as that missing conceptualisation, a theoretical construct whose characterisation framework is defined by two orthogonal distinctions: semantics $\leftrightarrow$ pragmatics, and properties $\leftrightarrow$  affordances, allowing to describe, compare, navigate, and potentially transform KGs across the full range of modelling practices. 
The methodological stance is empirical: rather than prescribing how KGs should be designed and built, the continuum aims to define a theory of the existent, derived from observation of real-world KG engineering practices and whose structure can be made formally explicit, for example, through Formal Concept Analysis (FCA). 
We ground the vision through a case study on provenance knowledge, showing how a single concern manifests differently across the continuum.
We articulate five open research challenges and invite the community to develop the ontological continuum as a shared research agenda.
\end{abstract}
\end{onlyextended}

\section{Introduction\ppinfo{1}}\begin{onlyshort} \vspace{-0.2cm}
\end{onlyshort}

\begin{inboth}
Knowledge graphs have become critical for data integration across scientific, governmental, and commercial domains. 
The diversity of KG modelling practices, from lightweight vocabularies to richly axiomatised ontologies, reflects the breadth of problems and communities they serve.
This diversity is an asset rather than a weakness: different communities\footnote{A community is intended here as a group of agents (human actors or organisations) with shared practices, expertise, and tasks: \emph{who} uses the knowledge and \emph{how.}} and tasks genuinely require different representational choices, thus creating a rich and varied KG ecosystem. 
The cost of this diversity, however, is restricted reuse. 
Adapting KG content across communities and tasks, e.g. repurposing a provenance model built for scientific workflows in an open data portal, or reusing ontological fragments defined under one foundational ontology in a setting grounded in another, remains expensive, brittle, and poorly understood.
On the one hand, neuro-symbolic methods bring novel tasks, which require the adaptation of existing KGs.
On the other hand, large language models offer the unprecedented capability to automate KG engineering at scale~\cite{garijo2024llms}. 
However, without a principled map of the space they are navigating, such automation remains conceptually ungrounded. 
\end{inboth}
\begin{onlyextended}
Emerging work on autonomous knowledge engineering agents~\cite{garijo2024llms} (systems that decompose, plan, and execute re-engineering decisions with minimal human intervention) makes this gap particularly acute: an agent navigating the KG ecosystem without a principled framework is forced to treat it as a black box, with no basis for choosing transformations, estimating their cost, or explaining its choices.
\end{onlyextended}

\begin{inboth}
%
%
The dominant response to KG diversity has been structural harmonisation, yet cross-KG reuse remains expensive and brittle in practice, and the premise deserves scrutiny.
We pose that the right question is not ``how similar is \textbf{KG\textsubscript{1}} to \textbf{KG\textsubscript{2}}?'' but rather: ``does \textbf{KG\textsubscript{1}} afford what community $C$ needs for task $T$? And if not, what would it take, and at what cost, to get there?'' 
We reframe reuse as a \emph{transformation problem}: given a source KG and a community with a task, what sequence of principled steps leads to a target that affords what is needed? 
Answering this question requires not a mapping technique but \textit{a map of the problem space} itself.
What is missing is a principled account of how diverse KG modelling practices interplay across both their semantics and their pragmatics.
We propose the \emph{ontological continuum} as that missing conceptualisation.

The contribution of this paper is \emph{conceptual} and \emph{terminological}. 
We define the \textit{ontological continuum}: a characterisation framework, organised around two orthogonal distinctions, \emph{semantics versus pragmatics}, and \emph{properties versus affordances}, that provides a vocabulary for describing, comparing, navigating, and potentially transforming KGs across the full range of modelling practices. 
In brief, the first distinction contrasts \emph{semantic} features, i.e.\ the meanings a KG encodes, with \emph{pragmatic} features, i.e.\ the mechanisms by which it is represented and operated on.
The second contrasts \emph{properties}, features intrinsic to the KG as an artefact, with \emph{affordances}, which are relational, constituted by the interplay of a KG, a community, and a task.
We ground the vision through a case study on provenance knowledge, 
showing how a single concern manifests differently across the continuum 
and how the structure of those positions can be made explicit and formally 
ordered through Formal Concept Analysis (FCA)~\cite{ganter1999formal,poelmans2013formal}. 
We articulate five open research challenges and discuss their implications.
\end{inboth}

\begin{onlyextended}
The paper is structured as follows. 
Section 2 surveys related work. Section 3 introduces the ontological continuum: first as a theoretical construct (Section 3.1), then as a characterisation framework (Section 3.2), and finally as an empirically-derived discrete projection (Section 3.3). Section 4 presents the provenance case study. 
Section 5 discusses implications. Section 6 articulates open research challenges. 
Section 7 concludes.
Details on the case studies, including tables and diagrams of the FCA, are provided in the appendices.
\end{onlyextended}

\section{Background and Related Work\ppinfo{starts at p3, 1.5}}\begin{onlyshort}\vspace{-0.2cm}
\end{onlyshort}
\label{sec:background}

\begin{inboth}
\begin{sloppypar}
The KG ecosystem spans from lightweight vocabularies (e.g. Dublin Core~\cite{dcterms}, Schema.org~\cite{schemaorg}) to richly axiomatised ontologies (e.g. Gene Ontology~\cite{gene2019gene}, SNOMED~CT~\cite{chang2021use}) and this diversity is both healthy and necessary: different communities have developed different formalisms, governance practices, and modelling methodologies, each fit for its own purpose. 
Yet diversity comes at a cost, and this is often not fully acknowledged. 
Knowledge and ontology engineering methodologies such as NeOn~\cite{suarez2011neon}, eXtreme Design~\cite{blomqvist2010experimenting}, LOT~\cite{poveda2022lot}, and MOMO~\cite{shimizu2023modular} provide principled guidance for constructing an ontology to meet a given set of requirements, and several, including competency-question-driven and modular design methods, actively support reuse.
What they do not offer is a principled account of characterising and navigating the diversity of existing KGs for cross-community reuse, which remains poorly understood and underspecified.
This diversity runs deeper than formalisms: foundational ontology choices, BFO~\cite{otte2022bfo}, DOLCE~\cite{borgo2022dolce}, UFO~\cite{guizzardi2022ufo}, create incompatibilities at the level of the conceptual model with real downstream consequences. 
Antipatterns defined within a UFO-based validation framework cannot be straightforwardly applied in a DOLCE-based setting; bridging them requires explicit mappings, operationalised as SPARQL and SHACL, a form of ad~hoc pragmatic adaptation for which currently there is no principled support~\cite{sales2015antipatterns,morlidge2026animlontology}.
\end{sloppypar}
\end{inboth}

\begin{inboth}
Classifying ontologies is itself a long-standing concern in knowledge engineering.
Ontologies have been sorted by their level of generality, into top-level, domain, task, and application ontologies~\cite{guarino1998formal,vanheijst1997using}, and characterised through their relationship to lexical and linguistic resources~\cite{huang2010ontology}.
A parallel tradition tied knowledge components to the tasks they serve, through task ontologies and reusable problem-solving methods~\cite{studer1998knowledge,chandrasekaran1999ontologies}.
These frameworks classify ontologies by type, content, or level of abstraction; the ontological continuum instead characterises a KG along task-relative dimensions that cut across this typology, combining what a KG means with what it affords, and its intrinsic properties with its use-relative possibilities.
However, like problem-solving methods made reasoning components task-specific, the continuum makes the characterisation of a KG task- and community-relative.
\end{inboth}

\begin{inboth}
The standard response to heterogeneity has been harmonisation: mapping languages (R2RML~\cite{das2012r2rml}), ontology alignment systems (LogMap~\cite{logma_ecai2012}, AML~\cite{faria2023AML}), and modularisation techniques~\cite{LeClair2022modularization} operationalise integration but fall short of addressing diversity in a unified methodology. 
Ontology alignment systems take full ontologies as
input and are evaluated against complete reference alignments~\cite{portisch2024background,hertling2023olala},
and evaluation metrics reward full coverage even if fit-for-purpose
reuse requires only the task-relevant fragment.
Pay-as-you-go engineering approaches~\cite{Sequeda2019pay} and ontology modularisation
have partially addressed this issue by focussing on ontology fragments, but without considering an affordance framing, both
remain task- and community-agnostic.
The ontological continuum framing makes the 
quality criterion task-relative coverage, rather than completeness.
\end{inboth}

\begin{inboth}A methodological precedent to the notion of continuum is Fa\c{c}ade-X~\cite{asprino2023knowledge}, which derives a unified access mechanism for heterogeneous non-RDF data sources by empirically analysing their structural characteristics and abstracting a minimal-commitment meta-model, enabling SPARQL querying without schema harmonisation. 
The stance is bottom-up and observation-first: the meta-model is abstracted from the structures data formats exhibit.\footnote{That this direction is gaining traction is signalled by the W3C Data Fa\c{c}ades Community Group~\cite{w3c:facadex}, which is actively developing standards in this space.} 
\end{inboth}
\begin{onlyextended}The ontological continuum adopts the same methodological stance, extending it from minimal-commitment access over heterogeneous data sources to the full space of KG modelling practices. 
\end{onlyextended}

\begin{inboth}
A growing body of work applies GenAI and LLMs to knowledge engineering tasks, ontology design, competency question engineering, SPARQL generation, and schema induction~\cite{garijo2024llms}, with important results. 
They share, however, a common gap: they operate without a principled model of the KG space they are navigating.
Ignoring this gap can be critical; the historical lesson of semantic web services is instructive: OWL-S, WSMO and related work modelled the semantic layer of service description well but left the pragmatic layer, i.e. how resources are composed and used in practice, unaddressed, and the initiative collapsed under that weight~\cite{cabral2004approaches,WSMO2006,owls2004,martin2004owls}. 
One might assume that GenAI now delivers the flexibility that semantic web services lacked; but without a principled pragmatic framework, the same limitations apply~\cite{payne-et-al2026}.
\end{inboth}


\begin{inboth}\begin{sloppypar}
Foundation literature on affordances, e.g. Gibson's ecological 
affordances~\cite{gibson1977theory} and Norman's design 
affordances~\cite{norman1999affordance}, has been extensively 
applied in HCI and design, but has not been systematically adopted 
in Knowledge and KG Engineering. Recently, Celino~\cite{celino2026ka} 
argued the utility of affordance-based characterisations within the 
information-seeking domain, where affordances capture what a knowledge 
source offers to an agent satisfying a query. We argue that in knowledge engineering, 
interoperability has a broader remit: it is constituted not only by the 
ability to answer a query but by the operations that a KG's 
formalisation, conceptual model, and representational choices make 
available to a community. No existing framework provides a principled account of how diverse KG modelling practices interplay at 
this level, leaving both neuro-symbolic integration and GenAI-driven 
automation without a map of the space they are navigating.
\end{sloppypar}
\end{inboth}

\section{The Ontological Continuum: A Vision\ppinfo{ends at p5: 1.5}}\begin{onlyshort}\vspace{-0.2cm}\end{onlyshort}

\begin{onlyshort}
We introduce the ontological continuum as a theoretical construct 
to drive transformations: a connected space of possible KG engineering solutions with no sharp divisions between them. 
What varies is a KG's ontological commitment, broadly construed: not only axiomatisation and alignment to foundational ontologies (e.g.\ BFO, DOLCE, UFO), but every design choice that determines what a KG can mean and do.
At one end lie lightweight vocabularies, usable with minimal domain knowledge; at the other, richly axiomatised KGs with explicit upper-ontology commitments that express more but require more to be used; between them, a continuous and densely populated range.
Positioning a single KG tells us where it sits; positioning a collection reveals the structure of the 
space between them: which transformations are incremental, 
which are costly, and which paths exist between a source and a 
target.

We use the term \emph{continuum} in a conceptual, not a 
mathematical, sense: the space is not given a priori but built empirically from observation of real-world practices, determined by the need to describe, compare, navigate, and transform KGs in principled ways. 
In particular, we do not claim a single linear ordering of KGs, such as a line from Dublin Core to PROV-O; a KG's position is a profile across several partial orders that need not coincide.
Positions are points in a space that may not be fully connected: most KGs can be connected, but paths that cross certain epistemic boundaries may not be available and require abandoning incremental transformation and choosing a different KG as the starting point.
\end{onlyshort}

\begin{onlyextended}
\subsection{The Continuum as a Theoretical Construct}
We introduce the ontological continuum as a theoretical construct: a connected space of possible KG engineering solutions with no sharp divisions between them. 
What varies across the continuum is the degree and nature of a KG's ontological commitment, understood in a broad sense that encompasses not only the degree of axiomatisation and alignment to foundational ontologies (e.g. BFO, DOLCE, UFO), but the full range of choices that determine what a KG can mean and what it can do. 
At one end lie KGs with minimal schema conventions and broad applicability: lightweight vocabularies that impose few constraints, carry few ontological assumptions, and can be queried and integrated with minimal domain knowledge. 
At the other end lie KGs with rich axiomatic structure, explicit upper-ontology commitments, and tightly scoped deployment contexts: they express more, but require more to be used. 
Between these extremes lies a continuous and densely populated range.
We use the term continuum in a conceptual, not a mathematical, sense. We do not posit an a priori infinite space of independently existing KG types. 
Rather, the continuum is a theoretical construct that we build empirically: it is derived from observation of real-world KG engineering practices, and it exists relative to an operational need, the need to describe, compare, navigate, and transform KGs in principled ways. 
What makes it a continuum is the claim that positions within it are not discrete types but points in a connected space: any two KGs, however different, can in principle be connected by a sequence of incremental transformations, each of which moves along one or more of its \textit{dimensions}. 
We introduce these dimensions in the next section.
The \textit{population} of the continuum is not given in advance; it is constructed from the properties and affordances that KGs actually exhibit, and it is therefore open to revision as practices evolve. 
This is a defining difference from prescriptive ontology methodologies: the continuum aims to be a \emph{descriptive theory}, account of what KG engineering practices are rather than rather than a normative account of what they should be.
\end{onlyextended}

\begin{onlyshort}
Describing a KG's position requires a vocabulary structured along two orthogonal distinctions. 
The \textbf{first distinction} separates \emph{semantic} from \emph{pragmatic} features. 
Semantic features concern the meanings a KG encodes: the domain knowledge it makes explicit, the concepts it commits to, the relations it asserts. 
Pragmatic features concern the mechanisms by which a KG is represented and operated on: the (meta-)models used to structure its statements (e.g.\ RDF, OWL, RDF-Star, the Wikidata model) and the interfaces through which it is queried, validated, or reasoned over. 
The \textbf{second distinction} separates \emph{properties} 
from \emph{affordances}. 
Properties are features intrinsic to the 
KG as an artefact, present regardless of who uses it or for what 
purpose. 
Affordances are relational: constituted by the 
relationship between a KG, a \emph{community}, and a 
\emph{task}, extending Gibson's ecological 
affordances~\cite{gibson1977theory} and Norman's design 
affordances~\cite{norman1999affordance} to KG engineering, where an affordance is not intrinsic to the KG but emerges from its relationship with a community pursuing a task.

We illustrate the two distinctions considering the case of provenance knowledge: \emph{semantic properties} (e.g.\ authorship, lineage, epistemic status), \emph{semantic affordances} (e.g.\ attribution, reproducibility), \emph{pragmatic properties} (e.g.\ named graphs, reification, PROV-O), and \emph{pragmatic affordances} (e.g.\ SPARQL, SHACL, OWL~DL reasoning).
\end{onlyshort}

\begin{onlyextended}
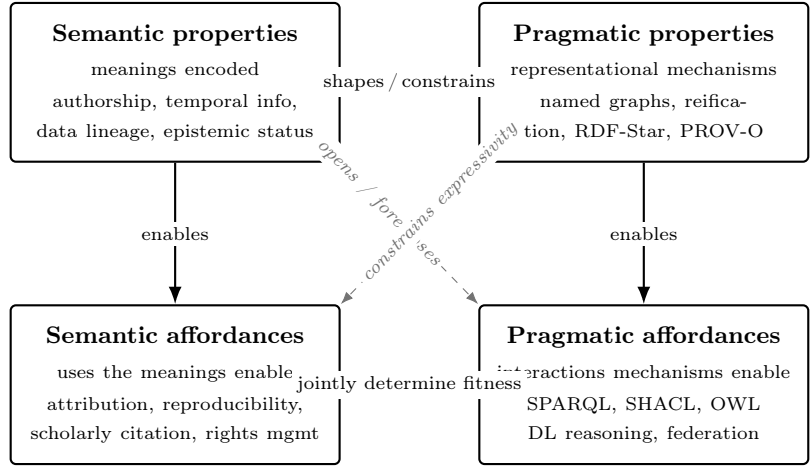
\begin{figure}[t]
\centering
\begin{tikzpicture}[
    font=\small,
    cell/.style={
        rectangle, draw, rounded corners=2pt, thick,
        minimum width=4.3cm, minimum height=2.1cm,
        align=center, inner sep=5pt, text width=4.0cm
    },
    solidlink/.style={-{Latex}, thick},
    bilink/.style={{Latex}-{Latex}, thick},
    diaglink/.style={-{Latex}, thin, dashed, black!55},
    elab/.style={font=\scriptsize, fill=white, inner sep=2pt},
    dlab/.style={font=\scriptsize\itshape, fill=white, inner sep=1.5pt, text=black!55}
]
\node[cell] (sp) at (-3.1, 2.0)
    {{\bfseries Semantic properties}\\[2pt]{\scriptsize meanings encoded}\\[1pt]
     {\scriptsize authorship, temporal info, data lineage, epistemic status}};
\node[cell] (pp) at ( 3.1, 2.0)
    {{\bfseries Pragmatic properties}\\[2pt]{\scriptsize representational mechanisms}\\[1pt]
     {\scriptsize named graphs, reification, RDF-Star, PROV-O}};
\node[cell] (sa) at (-3.1,-2.0)
    {{\bfseries Semantic affordances}\\[2pt]{\scriptsize uses the meanings enable}\\[1pt]
     {\scriptsize attribution, reproducibility, scholarly citation, rights mgmt}};
\node[cell] (pa) at ( 3.1,-2.0)
    {{\bfseries Pragmatic affordances}\\[2pt]{\scriptsize interactions mechanisms enable}\\[1pt]
     {\scriptsize SPARQL, SHACL, OWL DL reasoning, federation}};

\draw[diaglink] (sp.south east) -- (pa.north west)
    node[dlab, pos=0.28, sloped] {opens / forecloses};
\draw[diaglink] (pp.south west) -- (sa.north east)
    node[dlab, pos=0.28, sloped] {constrains expressivity};

\draw[solidlink] (sp.south) -- (sa.north) node[elab, midway] {enables};
\draw[solidlink] (pp.south) -- (pa.north) node[elab, midway] {enables};

\draw[bilink] (sp.east) -- (pp.west) node[elab, midway] {shapes\,/\,constrains};
\draw[bilink] (sa.east) -- (pa.west) node[elab, midway] {jointly determine fitness};
\end{tikzpicture}
\caption{The four characterisation dimensions as a 2$\times$2 of two
orthogonal distinctions: \emph{semantic}\,/\,\emph{pragmatic} (columns)
and \emph{property}\,/\,\emph{affordance} (rows). Solid vertical arrows:
properties \emph{enable} affordances, relative to a community pursuing a
task. Solid horizontal arrows: the two kinds of commitment interact at
each level. Dashed diagonals: cross-dimension dependencies
(Section~\ref{sec:discussion}, Challenge~2). Examples are drawn from the
provenance case study.}
\label{fig:four-dimensions}
\end{figure}
\subsection{Dimensions of the Continuum}
Describing a KG's position in the continuum requires a vocabulary structured along four characterisation dimensions, defined by two orthogonal distinctions, each capturing a fundamental partition in what we can say about a KG.

The first distinction separates \emph{semantic} from \emph{pragmatic} features. 
Semantic features concern the meanings that a KG encodes, i.e. the domain knowledge it makes explicit, the concepts it commits to, the relations it asserts. 
Pragmatic features concern the mechanisms by which a KG is represented and operated on, i.e. the meta-models used to structure its statements, the interfaces through which it is queried, validated, or reasoned over.

The second distinction separates \textit{properties} from \textit{affordances}. 
Properties are features intrinsic to the KG as an artefact: they are present regardless of who uses the KG or for what purpose. 
Affordances, by contrast, are relational: they are constituted by the relationship between the KG, a \textit{community}, and a \textit{task}. 
Formally, an affordance is a three-place predicate: \textit{affords}$(KG, community, task)$, that holds when a KG makes a particular kind of use possible for a particular community pursuing a particular task. 
This framing draws on the notion of affordance, developed in ecological psychology (Gibson~\cite{gibson1977theory}) and design theory (Norman~\cite{norman1999affordance}), as a relational property of an agent–environment pair, and extends it to the context of KG engineering: an affordance is not a feature of the KG alone, nor of the community alone, but of their meeting. 
Crucially, the same KG may afford different things to different communities, and characterising it fully requires naming not just its intrinsic properties but its \textit{use-relative possibilities}.
The two distinctions together define four characterisation dimensions, which we introduce.

\emph{Semantic properties} capture the meanings encoded by a KG: the factual, assertional, or inferential content that constitutes its knowledge about a domain. 
For example, in the context of provenance, semantic properties include authorship, temporal information, data lineage, epistemic status, and agent roles.

\emph{Semantic affordances} capture what a KG's semantic content makes possible for specific communities pursuing specific tasks; what downstream uses the encoded meanings support, relative to the agents who can interpret and exploit them. 
These are not features of the KG in isolation but of the KG relative to its interpreters: attribution, reproducibility, scholarly citation, and rights management are semantic affordances in the provenance context, enabled by the semantic properties a KG exhibits, but affordances only relative to communities that can and do exploit them.

\emph{Pragmatic properties} capture the representational mechanisms used to structure the KG's statements: the meta-models (RDF, OWL, RDF-Star), the Wikidata statement model, OAI-ORE aggregation, PROV-O, that determine how knowledge is encoded and made accessible. 
Pragmatic properties include the representational choices that constrain expressivity and interoperability: the adoption of named graphs, reification strategies, schema languages, and the degree of alignment to standard vocabularies.

\begin{sloppypar}\emph{Pragmatic affordances} capture the interaction methods that the KG's representational mechanisms make available to communities: SPARQL querying, SHACL constraint validation, OWL DL reasoning, bulk download, REST API access, SPARQL federation. 
Like semantic affordances, these are relational: SPARQL federation is only an affordance relative to a community that has the infrastructure and expertise to exploit it; OWL DL reasoning is only an affordance relative to a task that requires it.\end{sloppypar}
\end{onlyextended}

\begin{inboth}
Each feature's dimension is determined by two questions, not by convention: whether it is semantic or pragmatic, and whether it is a property or an affordance.
A feature is \emph{semantic} if it concerns what the KG says about a domain, and \emph{pragmatic} if it concerns how the KG is encoded and operated on.
It is a \emph{property} if it can be read off the artefact in isolation, and an \emph{affordance} if realising it requires positing an agent and a task.
Named graphs, reification, RDF-Star, and PROV-O are representational mechanisms materialised in the data, present regardless of use, and are therefore pragmatic \emph{properties}.
SPARQL querying, SHACL validation, and OWL DL reasoning are operations an agent performs, realised only when a community with the requisite tooling engages the KG for a task, and are therefore pragmatic \emph{affordances}.
The same representational choice may be a property and at the same time ground affordances: PROV-O is a pragmatic property that enables OWL reasoning as a pragmatic affordance.
Such cross-dimension co-occurrence is expected, not a flaw in the framework.
Indeed, a mature continuum theory may reveal a systematic dependency structure between properties and affordances.
\end{inboth}

\begin{onlyshort}
\begin{table}[t]
\centering
\small
\caption{The four characterisation dimensions, with selected provenance features for each (42 in total across the ten KGs); full feature matrices and FCA lattices appear in the extended version~\cite{continuumex}.}\vspace{-0.2cm}
\label{tab:provenance-summary}
\begin{tabular}{@{}l l l@{}}
\toprule
 & \textbf{Properties} & \textbf{Affordances} \\
 & \footnotesize\textit{(intrinsic to the KG)} & \footnotesize\textit{(relative to community + task)} \\
\midrule
\textbf{Semantic}  & authorship, temporal information,   & attribution, reproducibility, \\
\footnotesize\textit{(meanings)}
                   & data lineage, epistemic status,     & scholarly citation, rights \\
                   & agent roles \emph{(14 total)}       & management \emph{(10 total)} \\
\addlinespace
\textbf{Pragmatic} & named graphs, RDF reification,      & SPARQL, SHACL validation, \\
\footnotesize\textit{(mechanisms)}
                   & RDF-Star, Wikidata model,           & OWL DL reasoning, SPARQL \\
                   & PROV-O \emph{(7 total)}             & federation \emph{(11 total)} \\
\bottomrule
\end{tabular}\vspace{{-0.3cm}}
\end{table}
\end{onlyshort}

\begin{onlyshort}
Any operational analysis of the continuum requires a projection onto 
a finite, discrete structure. Given the four dimensions defined, 
the features of each KG can be recorded per dimension, producing a 
binary feature matrix: objects (KGs) crossed with attributes 
(dimension features). Formal Concept Analysis 
(FCA)~\cite{ganter1999formal} provides a formal method for 
projecting these matrices into an ordered space: a 
concept lattice in which each formal concept is defined by a 
maximal set of objects sharing a common attribute set. 
In our setting, each dimension yields a lattice that orders KGs by the richness of their characterisation profile (similarly, a lattice with all integrated dimension features can be built): towards one end of the lattice, more KGs share a few features; towards the other end, more features will be shared by few KGs.
\end{onlyshort}

\begin{onlyextended}
\subsection{Discrete Projection in the Continuum}
The ontological continuum is a theoretical construct; any operational analysis of such continuum needs to necessarily be a projection onto a finite, discrete structure. 
With the four dimensions defined, we can analyse the features of each source KG as well as the target, desired KG, to produce a list, organised under those dimensions.
The result would be a binary feature matrix for each of the four dimensions: objects (KGs) crossed with attributes (dimension features). 

Formal Concept Analysis (FCA) \cite{ganter1999formal} provides a principled method for projecting these features into a mathematically ordered space.
FCA is a 
method for deriving a conceptual hierarchy from a binary table relating 
objects to attributes. 
Such a table, called a \emph{formal context}, 
records which objects possess which attributes; from it, FCA computes a 
set of \emph{formal concepts}, each a maximal pairing of a set of objects 
with the set of attributes they all share. These concepts are partially 
ordered by inclusion and organised into a \emph{concept lattice}, in which 
more specific concepts (many shared attributes, few objects) sit below 
more general ones (few shared attributes, many objects). The lattice makes 
the implicit structure of the data explicit and navigable, which is why we 
adopt it to order KGs by the features they exhibit.

In our setting, each dimension yields a lattice that orders KGs by the richness of their characterisation profile: KGs at the bottom of the lattice share the densest, most specific attribute sets; those toward the top are characterised by fewer, more general features. 
Similarly, we can produce a unified lattice, by grouping all features together.

\end{onlyextended}

\begin{onlyshort}\vspace{-0.2cm}
\end{onlyshort}
\section{Case Study: Provenance Across the Continuum\ppinfo{Starts at page 6, 0.5 the method + 0.5p each case study, 1.5}}\label{sec:provenance}\begin{onlyshort}\vspace{-0.2cm}\end{onlyshort}

\begin{onlyshort}
We instantiate the characterisation framework on provenance 
knowledge across ten KGs (see 
Table~\ref{tab:provenance-summary}).\footnote{A common, simpler task would only consider a source KG and a target one. To show the generality of the approach, we analyse ten KGs.}
Provenance representations are diverse along all four 
characterisation dimensions, from a single lightweight 
property (\texttt{dc:source}) to a full ontological model 
of agents, activities, and entities in PROV-O.
\end{onlyshort}

\begin{onlyextended}
We present a case study on 
\emph{provenance knowledge} to demonstrate that the 
characterisation framework can be instantiated concretely, i.e. 
that existing KGs can be positioned within the 
continuum and that the resulting structure is principled 
and interpretable. 
The representation of provenance provides an ideal use-case: 
representations are genuinely diverse along all four 
dimensions, from a single lightweight property 
(\texttt{dc:source}) to a full ontological model of 
agents, activities, and entities in PROV-O.
\end{onlyextended}

\begin{inboth}
We perform the analysis with the aid of a generative AI 
model (Claude Sonnet 4.6), supervising each step: features 
are assigned per dimension, existing 
features are reused wherever possible, and any new feature 
introduced is verified retrospectively against previously 
analysed KGs to ensure terminological consistency. 
The model proposed candidate features for each KG and dimension from the relevant documentation and vocabularies; the authors reviewed every proposal, and each assignment recorded in the matrices was author-verified rather than accepted automatically\footnote{The prompts used, the per-KG feature assignments, and the scripts' generation prompts are all available at \url{https://claude.ai/share/4e88685f-6a20-41d1-9939-c120c505de5c} for inspection. Data and scripts are shared alongside the paper latex source code as supplementary material~\cite{supplementary}}.

The result is a binary feature matrix for each dimension: 
objects (KGs) crossed with attributes (dimension+features), from which a concept lattice is derived via 
FCA~\cite{ganter1999formal}. 
Artefacts sharing the densest, most specific attribute sets cluster at the 
bottom of the lattice; those characterised by fewer, more 
general features appear towards the top.
\end{inboth}

\begin{onlyextended}
The formal context produced, i.e. the choice of objects, 
attributes, and the incidence relation between them, is 
relative to the community and task that guide the 
analysis: a different community, or a different task, 
would select different attributes, produce a different 
formal context, and derive a different though equally 
principled concept lattice. The dependencies between 
attributes across dimensions (which semantic properties 
entail which pragmatic affordances, and which pragmatic 
choices foreclose which semantic commitments) are not yet 
formalised; making them explicit is the substance of 
Challenge~2 (Section~\ref{sec:challenges}).
\end{onlyextended}

\begin{onlyextended}
We analyse ten KGs, each representing a distinct approach 
to provenance.

\textbf{Europeana} uses the Europeana Data Model (EDM), 
capturing provenance through OAI-ORE Aggregations that 
record the original data provider, intermediaries, and 
Europeana as aggregator, with Dublin Core metadata at the 
core~\cite{haslhofer2011}.

\textbf{Google Data Commons} extends Schema.org to attach 
a provenance property to every triple, recording the 
originating provider's URL as a first-class element of the 
data model~\cite{datacommons,schemaorg}.

\textbf{Bio2RDF} stores per-dataset provenance in 
dedicated named graphs, using VoID, PROV, and Dublin Core 
to describe conversion dates, licensing, and generating 
scripts~\cite{callahan2013}.

\textbf{The British Museum ResearchSpace} platform uses 
RDF reification grounded in CIDOC-CRM to capture the 
provenance of scholarly assertions about cultural heritage 
objects~\cite{researchspace}.

\textbf{UniProt} has adopted RDF-Star to attach evidence 
and provenance annotations to individual triples, 
replacing a verbose and costly reification-based 
approach~\cite{rdfstar,uniprot}.

\textbf{Wikidata} provides fine-grained triple-level 
provenance through a bespoke statement model of 
qualifiers and references~\cite{vrandevcic2014,hernandez2015}.

\textbf{The EU Open Data Portal} captures dataset 
provenance through DCAT-AP, incorporating PROV-O via 
\texttt{prov:qualifiedAttribution} to express agent roles 
and data lineage~\cite{provo,dcatap}.

\textbf{DBpedia} organises Wikipedia-extracted data into 
named graphs per language and dataset, using PROV-O to 
record extraction process and originating 
source~\cite{dbpedia}.

\textbf{Linked Open Vocabularies (LOV)} tracks vocabulary 
publication provenance using PROV-O, recording authors, 
publishers, versioning history, and curation 
agents~\cite{lov}.

\textbf{Nanopublications} treat provenance as a structural 
concern: each nanopublication comprises three named graphs:  assertion, provenance, and publication info, with 
PROV-O capturing authorship, temporal data, and the 
epistemic status of scientific claims~\cite{nanopub}.
\end{onlyextended}

\begin{inboth}
\begin{sloppypar} 
The ten KGs span the continuum along all four dimensions (see Table~\ref{tab:provenance-combined}).
The feature vocabulary is derived empirically during the analysis, reusing terms across KGs for consistency, not from a pre-existing controlled vocabulary.
Semantic content ranges from dataset-level attribution and source URLs to fine-grained authorship, lineage, and epistemic status at the assertion level, and the affordances it enables differ accordingly: the EU Open Data Portal's PROV-O-based DCAT-AP affords regulatory compliance tracking that a \texttt{dc:source} annotation does not, while Nanopublications and UniProt support data quality assessment.
Pragmatic mechanisms range from simple Dublin Core annotations to PROV-O's full agent--activity--entity model; all ten KGs are published in RDF, but it is the richer commitments layered on top (OWL vocabularies such as PROV-O and CIDOC-CRM, or RDF-Star and the Wikidata statement model) rather than RDF alone, that determine which reasoning and validation affordances (e.g.\ OWL DL reasoning, SHACL) are available.

We conjecture that the two kinds of feature generalise differently: semantic features are relative to the subject domain, here provenance, yet tend to recur across the communities that share it, whereas pragmatic features are largely subject-domain-independent standards whose selection is nonetheless specific to a KG's ecosystem.
A general account of which features are domain-independent, and its implications for reuse, is left to future work.\end{sloppypar}
\end{inboth}
\begin{onlyshort}
A full account of the process, the binary feature matrices, and the FCA lattices for the four dimensions are provided in the extended version~\cite{continuumex}.
\end{onlyshort}
\begin{onlyextended}
The complete feature matrices for the four dimensions are given in the
appendix (Table~\ref{tab:provenance-combined} consolidates them; the
per-dimension matrices appear in the appendix tables), and the four
concept lattices derived from them are shown in
Figures~\ref{fig:fca-semantic_properties}--\ref{fig:fca-pragmatic_affordances},
each accompanied by a legend
(Tables~\ref{tab:fca-semantic_properties}--\ref{tab:fca-pragmatic_affordances})
listing, for every concept, the KGs it covers and the features they share.
We read each lattice top-down, from the most general concept (few shared
features, many KGs) to the most specific (many shared features, few KGs).

\emph{Semantic properties} (Figure~\ref{fig:fca-semantic_properties},
Table~\ref{tab:fca-semantic_properties}). No semantic property is shared by
all ten KGs: the top concept has an empty intent, confirming that what a KG
records \emph{about} provenance is the most variable dimension. The first
splits are the recurring coarse descriptors --- \emph{data provider},
\emph{authorship}, and \emph{temporal information}, each shared by seven KGs
--- while the discriminating depth comes from \emph{triple-level lineage}
(Google Data Commons, UniProt, Wikidata, Nanopublications) and the cluster
of \emph{epistemic status}, \emph{evidence type}, and \emph{scholarly
assertion} that pulls the assertion-level KGs (Wikidata, UniProt,
Nanopublications, British Museum ResearchSpace) towards the bottom. Europeana
sits apart, the only KG carrying \emph{aggregation chain}.

\emph{Semantic affordances} (Figure~\ref{fig:fca-semantic_affordances},
Table~\ref{tab:fca-semantic_affordances}). Here the top concept is non-empty:
\emph{attribution} is afforded by every KG, the one universal semantic
affordance. Below it the lattice separates the data-publishing affordances
(\emph{data discovery}, \emph{rights management}, \emph{cross-portal
interoperability}, clustering Europeana, EU ODP, LOV, DBpedia) from the
scientific affordances (\emph{reproducibility}, \emph{data quality
evaluation}, \emph{scholarly citation}, \emph{knowledge curation}, clustering
UniProt, Wikidata, Nanopublications). The bottom concepts are the richest
profiles: Wikidata and Nanopublications jointly afford citation, curation,
and quality evaluation, the densest semantic-affordance signature in the set.

\emph{Pragmatic properties} (Figure~\ref{fig:fca-pragmatic_properties},
Table~\ref{tab:fca-pragmatic_properties}). This is the sparsest and most
clearly separated lattice: most KGs sit on a single mechanism, so the
concepts are almost disjoint. Six KGs occupy their own atoms ---
\emph{OAI-ORE} (Europeana), \emph{Schema.org markup} (Google Data Commons),
\emph{RDF reification} (British Museum ResearchSpace), \emph{RDF-Star}
(UniProt), and the \emph{Wikidata model} (Wikidata) --- while \emph{PROV-O}
is the one shared mechanism (EU ODP, DBpedia, LOV, Nanopublications), and
\emph{named graphs} co-occurs with it for DBpedia and Nanopublications. The
near-disjointness is itself the finding: the choice of provenance mechanism
is where KGs diverge most decisively.

\emph{Pragmatic affordances} (Figure~\ref{fig:fca-pragmatic_affordances},
Table~\ref{tab:fca-pragmatic_affordances}). The pattern inverts: this is the
densest lattice, with \emph{SPARQL} afforded universally (top concept) and a
long chain of widely shared affordances --- \emph{SPARQL endpoint},
\emph{bulk download}, \emph{linked data browsing}, \emph{REST API} ---
beneath it. Discrimination comes only at the specialised end:
\emph{OWL DL reasoning} (British Museum ResearchSpace, EU ODP,
Nanopublications), \emph{SHACL} (Wikidata, EU ODP), \emph{SPARQL federation}
(UniProt, Wikidata, DBpedia), and the singletons \emph{SPARQL-Star} (UniProt)
and \emph{natural language query} (Google Data Commons).

Read together, the four lattices make the central claim of the continuum
concrete: a KG's position is not a single point but a profile across four
orders that do not coincide. Wikidata sits low (most specific) in the
semantic and the affordance lattices yet on an isolated atom in the
pragmatic-properties lattice; the EU Open Data Portal is rich in pragmatic
affordances and PROV-O-based properties but mid-depth in semantics. 
These distinctions that locate a KG in the continuum and will determine which transformations separate it from each other.
\end{onlyextended}
\begin{onlyextended}
\section{Discussion}\label{sec:discussion}
We discuss implications and lessons learnt from the ontological continuum case study exercise.
A lattice itself is not \textit{the continuum}: it is a discrete projection of it, sampling a finite population of KGs with binary feature judgements. 
It approximates the structure of the continuum in a principled way: the partial order reflects the actual co-occurrence structure of characterisation features across real-world KGs, derived bottom-up from observation rather than imposed top-down from a prior taxonomy. 
Other projection methods are possible and may be preferable for different tasks; FCA is our choice here because it makes the ordering explicit and interpretable.

Two properties of the projection are worth stating explicitly. 
First, the projection is finite: the lattices reflect only the KGs analysed and the features they exhibit; new KGs or new features may alter the structure, and the methodology is designed to accommodate iterative extension. 
Second, the projection is task-relative: the lattice produced for a provenance-focused analysis differs from one produced for bioinformatics or cultural heritage metadata, and different community practices give rise to different feature distributions. 
The continuum itself is intended as a task- and community-independent theoretical construct; its discrete projections may still be task-specific instantiations. 
This is a feature of the approach, not a limitation: it means the framework can be instantiated for any community by following the empirical procedure described in Section 4, producing a lattice that reflects the actual structure of that domain's KG engineering practices rather than an abstract (or shared) ideal.

These four dimensions are interdependent.
Semantic commitments shape which pragmatic affordances are available: a KG that models provenance using PROV-O opens OWL reasoning as a pragmatic affordance while foreclosing some SPARQL-Star interaction patterns. 
Conversely, pragmatic choices constrain what semantic properties can be expressed: the adoption of the Wikidata statement model enables fine-grained triple-level provenance but embeds it in a bespoke meta-model not directly interoperable with PROV-O. 
One of the goals of the continuum is to make these cross-dimension dependencies explicit; formalising their directionality and transitivity is itself a core research challenge (Challenge 2, Section~\ref{sec:challenges}). A key ambition of the framework is to replace the informal notion of \textit{``community suitability''} with an operational one: rather than saying a KG is fit for a given community, we can say it exhibits a specific combination of semantic and pragmatic properties and affords a specific set of uses, thus turning fitness for purpose from an assumed judgement into an explicit, checkable combination of properties and affordances; which combination a given task requires remains an empirical question, not one the clustering settles by itself. 

The continuum is not intended to make the reuse decision itself: whether a community adopts a given KG may rest on factors beyond its semantic and pragmatic characterisation, including perceptions, preferences, and trust.
Its purpose is rather to operationalise re-engineering once that decision is made, characterising systematically the distinctions between a source and a target KG and the transformation that connects them.
The continuum may inform a reuse decision, but its remit is to make its realisation systematic.
\end{onlyextended}


\begin{onlyshort}\vspace{-0.2cm}
\end{onlyshort}
\section{\iftoggle{extended}{Research Agenda}{Discussion and Outlook}}\label{sec:challenges}\begin{onlyshort}\vspace{-0.2cm}
\end{onlyshort}

\begin{onlyshort}
The exemplary case study helps clarifying the framework's scope.
First, the pragmatic dimension deliberately spans not only representation but the \emph{access surface} through which a KG is published: bulk download, linked data browsing, and multilingual access are pragmatic affordances, because fitness for a task depends on how a KG is made available, not only on how it is encoded.
The same test keeps these affordances distinct from their underlying properties: \emph{multilinguality}, the multilingual content a graph carries, is a semantic property, whereas \emph{multilingual access}, the ability to retrieve that content by language, is a pragmatic affordance.
Second, structural and topological features, such as size, coverage, and level of detail, are intrinsic properties admitted by the same test; the provenance case study foregrounds representational properties, but the dimension is open to them.
Third, because affordances are relational and features are assigned by an analyst, applying the framework carries a degree of subjectivity; this is mitigated by recording each analysis as an explicit formal context, which makes every feature judgement inspectable and open to revision.    
\end{onlyshort}

\begin{onlyshort}
As a finite, domain-relative projection, the FCA lattice samples a specific collection of KGs with binary feature judgements. 
Different tasks naturally yield different feature distributions: a lattice for provenance integration differs from one for scientific workflow or cultural heritage metadata.
The four dimensions are interdependent: modelling provenance with PROV-O, for instance, opens OWL reasoning while foreclosing some SPARQL-Star interaction patterns.
A key ambition is to make these dependencies explicit, replacing the informal notion of \emph{community suitability} with an operational one: rather than calling a KG fit for a community, we say it exhibits specific properties and affords specific uses, turning \emph{fitness for purpose} from an assumed judgement into an explicit, checkable combination, though which combination a given task requires remains an empirical question, not one the feature clustering settles on its own.

The continuum may help evaluating a reuse decision based on non-functional needs: whether a community adopts a given KG may rest on factors beyond its semantic and pragmatic characterisation, including perceptions, preferences, and trust.
However, its purpose is rather to operationalise re-engineering once that decision is made, characterising systematically the distinctions between a source and a target KG and the transformation that connects them.
\end{onlyshort}

\noindent We propose the following challenges for the community:

\medskip\noindent{\textbf{Challenge 1: Characterising and 
formalising the dimensions}.}
The concrete semantic and pragmatic properties and 
affordances that populate the four characterisation 
dimensions must be derived empirically from real-world KGs 
and community practices, with community-relative fitness 
metrics defined in terms of the attribute combinations a 
KG must exhibit for a given task.
However, we conjecture that such list of features is finite.\begin{onlyshort}\footnote{Features are task- and community-relative projections onto a bounded vocabulary, so each application yields a finite set even if the theoretical total is open; which of them are domain-independent, and which domain- or ecosystem-specific, remains an open question.}
\end{onlyshort}
\begin{onlyextended}
Features are community- and task-relative projections onto a bounded vocabulary, so any given application generates a finite feature set even if the total theoretical set is open. 
A related open question is which of these features are domain-independent, and therefore reusable across subject domains and communities, and which are specific to a domain or ecosystem; the semantic--pragmatic distinction suggests a starting conjecture, but a general account remains open.
\end{onlyextended}

\smallskip\noindent{\textbf{Challenge 2: Formalising the 
dependencies across the continuum}.}
Semantic commitments shape which pragmatic affordances are 
available, and pragmatic choices constrain what semantic 
properties can be expressed; we hypothesise that making these dependencies 
explicit (including how upper-level ontology design choices cascade into pragmatic affordances) is essential for guided transformations.

\smallskip\noindent{\textbf{Challenge 3: Enabling modular, 
context-aware reuse}.}
Reuse requires extracting and applying fragments relevant 
to a given task from KGs that may differ substantially in 
ontological commitment, with explicit cost models for the adaptation involved. We hypothesise that the continuum theory may support developing reengineering metrics and cost models.

\smallskip\noindent{\textbf{Challenge 4: Supporting the 
co-evolution of interdependent KG engineering 
artefacts}.}
Changes in ontological commitment propagate into mappings, 
queries, and documentation; the continuum provides the 
vocabulary to track these effects, but tools are needed to 
act on them before commitments are made, and support KG management practices as a whole.

\smallskip\noindent{\textbf{Challenge 5: Developing 
human-in-the-loop strategies for users with varying 
expertise}.}
GenAI can bridge the gap between the continuum as an 
expert framework and the needs of non-expert users, but 
only if it can navigate the continuum in a principled, 
explainable way --- grounding transformation decisions in 
continuum dimensions, making costs explicit, and tailoring interactions to users with diverse expertise.

\iftoggle{extended}{\section{Conclusion}}{\medskip\noindent}

\begin{inboth}
The ontological continuum is the missing conceptualisation for principled KG reengineering: a vocabulary organised around two distinctions (semantics versus pragmatics, and properties versus affordances) that provides a backbone theory for describing, comparing, navigating, and transforming KGs across the full range of modelling practices.
The timing is right: GenAI offers the automation capability, neuro-symbolic AI creates the demand, and the diverse KG ecosystem on the web provides, for the first time, the empirical basis on which such a theory can be built.
We invite the community to pursue this shared research agenda.
\end{inboth}
\iftoggle{extended}{%
  \begin{table}[t]%
}{%
  \begin{table}[H]%
}
\centering
\scriptsize
\setlength{\tabcolsep}{4pt}
\renewcommand{\arraystretch}{1.2}
\caption{Provenance features of the ten knowledge graphs across the four characterisation dimensions. Each cell lists the features exhibited by that KG in that dimension. Generated from the same data as the per-dimension matrices.}
\label{tab:provenance-combined}
\begin{tabular}{>{\raggedright\arraybackslash}p{0.10\textwidth} >{\raggedright\arraybackslash}p{0.22\textwidth} >{\raggedright\arraybackslash}p{0.22\textwidth} >{\raggedright\arraybackslash}p{0.12\textwidth} >{\raggedright\arraybackslash}p{0.22\textwidth}}
\toprule
\textbf{KG} & \textbf{Semantic property} & \textbf{Semantic affordance} & \textbf{Pragmatic property} & \textbf{Pragmatic affordance} \\
\midrule
Europeana & authorship, data provider, aggregation chain, licensing, dataset-level lineage & attribution, data discovery, rights management, cross-portal interoperability & OAI-ORE aggregation & SPARQL, SPARQL endpoint, REST API, linked data browsing, bulk download, multilingual access \\
\addlinespace
Google Data Commons & data provider, source organisation, dataset-level lineage, triple-level lineage & attribution, data discovery, source tracking, cross-dataset integration & Schema.org markup & SPARQL, REST API, natural language query, bulk download \\
\addlinespace
Bio2RDF & data provider, dataset-level lineage, conversion process, licensing, temporal information & attribution, data discovery, source tracking, rights management, reproducibility & named graphs & SPARQL, SPARQL endpoint, linked data browsing, bulk download \\
\addlinespace
British Museum ResearchSpace & authorship, scholarly assertion, epistemic status, temporal information & attribution, source tracking, scholarly citation, knowledge curation & RDF reification & SPARQL, SPARQL endpoint, OWL DL reasoning, REST API, linked data browsing \\
\addlinespace
UniProt & authorship, evidence type, epistemic status, triple-level lineage, temporal information & attribution, source tracking, data quality evaluation, reproducibility & RDF-Star & SPARQL, SPARQL endpoint, SPARQL-Star, SPARQL federation, REST API, bulk download \\
\addlinespace
Wikidata & authorship, data provider, evidence type, epistemic status, triple-level lineage, temporal information & attribution, source tracking, data quality evaluation, scholarly citation, knowledge curation, cross-dataset integration & Wikidata model & SPARQL, SPARQL endpoint, SPARQL federation, SHACL, REST API, linked data browsing, bulk download, multilingual access \\
\addlinespace
EU ODP & data provider, authorship, dataset-level lineage, agent roles, temporal information & attribution, data discovery, rights management, reproducibility, cross-portal interoperability & PROV-O & SPARQL, SPARQL endpoint, OWL DL reasoning, SHACL, REST API, linked data browsing, bulk download, multilingual access \\
\addlinespace
DBpedia & data provider, dataset-level lineage, conversion process, multilinguality, licensing & attribution, data discovery, source tracking, cross-dataset integration, rights management & named graphs, PROV-O & SPARQL, SPARQL endpoint, SPARQL federation, REST API, linked data browsing, bulk download, multilingual access \\
\addlinespace
LOV & data provider, authorship, temporal information, dataset-level lineage, agent roles, licensing & attribution, data discovery, source tracking, cross-portal interoperability, knowledge curation & PROV-O & SPARQL, SPARQL endpoint, REST API, linked data browsing, bulk download \\
\addlinespace
Nanopubli-cations & authorship, temporal information, epistemic status, evidence type, scholarly assertion, triple-level lineage, agent roles & attribution, scholarly citation, knowledge curation, reproducibility, data quality evaluation, source tracking & named graphs, PROV-O & SPARQL, SPARQL endpoint, OWL DL reasoning, linked data browsing, bulk download \\
\addlinespace
\bottomrule
\end{tabular}
\end{table}




%
%
%
\bibliographystyle{splncs04}
\bibliography{main}
\begin{onlyextended}
\clearpage
\section*{Appendix: Provenance in the Continuum}

This appendix provides the full output of the provenance case study
(Section~\ref{sec:discussion}). The per-dimension feature matrices record,
for each of the ten KGs, which features it exhibits along each dimension:
semantic properties (Table~\ref{tab:semantic_properties}), semantic
affordances (Table~\ref{tab:semantic_affordances}), pragmatic properties
(Table~\ref{tab:pragmatic_properties}), and pragmatic affordances
(Table~\ref{tab:pragmatic_affordances}); Table~\ref{tab:provenance-combined}
consolidates all four into a single per-KG profile. From each matrix we
derive a concept lattice via FCA: semantic properties
(Figure~\ref{fig:fca-semantic_properties}), semantic affordances
(Figure~\ref{fig:fca-semantic_affordances}), pragmatic properties
(Figure~\ref{fig:fca-pragmatic_properties}), and pragmatic affordances
(Figure~\ref{fig:fca-pragmatic_affordances}). Each lattice is accompanied by
a legend (Tables~\ref{tab:fca-semantic_properties},
\ref{tab:fca-semantic_affordances}, \ref{tab:fca-pragmatic_properties},
and~\ref{tab:fca-pragmatic_affordances}) that lists, for every concept, the
set of KGs it covers and the features they share.
We do not produce a single FCA combining all dimensions.
\begin{table}[H]
\centering
\small
\begin{tabular}{lcccccccccc}
\toprule
 & \rotatebox{90}{\textbf{Europeana}} & \rotatebox{90}{\textbf{Google Data Commons}} & \rotatebox{90}{\textbf{Bio2RDF}} & \rotatebox{90}{\textbf{British Museum ResearchSpace}} & \rotatebox{90}{\textbf{UniProt}} & \rotatebox{90}{\textbf{Wikidata}} & \rotatebox{90}{\textbf{EU ODP}} & \rotatebox{90}{\textbf{DBpedia}} & \rotatebox{90}{\textbf{LOV}} & \rotatebox{90}{\textbf{Nanopublications}} \\
\midrule
\textit{authorship} & $\times$ &  &  & $\times$ & $\times$ & $\times$ & $\times$ &  & $\times$ & $\times$ \\
\textit{data provider} & $\times$ & $\times$ & $\times$ &  &  & $\times$ & $\times$ & $\times$ & $\times$ &  \\
\textit{temporal information} &  &  & $\times$ & $\times$ & $\times$ & $\times$ & $\times$ &  & $\times$ & $\times$ \\
\textit{dataset-level lineage} & $\times$ & $\times$ & $\times$ &  &  &  & $\times$ & $\times$ & $\times$ &  \\
\textit{licensing} & $\times$ &  & $\times$ &  &  &  &  & $\times$ & $\times$ &  \\
\textit{epistemic status} &  &  &  & $\times$ & $\times$ & $\times$ &  &  &  & $\times$ \\
\textit{triple-level lineage} &  & $\times$ &  &  & $\times$ & $\times$ &  &  &  & $\times$ \\
\textit{agent roles} &  &  &  &  &  &  & $\times$ &  & $\times$ & $\times$ \\
\textit{evidence type} &  &  &  &  & $\times$ & $\times$ &  &  &  & $\times$ \\
\textit{conversion process} &  &  & $\times$ &  &  &  &  & $\times$ &  &  \\
\textit{scholarly assertion} &  &  &  & $\times$ &  &  &  &  &  & $\times$ \\
\textit{source organisation} &  & $\times$ &  &  &  &  &  &  &  &  \\
\textit{multilinguality} &  &  &  &  &  &  &  & $\times$ &  &  \\
\textit{aggregation chain} & $\times$ &  &  &  &  &  &  &  &  &  \\
\bottomrule
\end{tabular}
\caption{Semantic properties of provenance across knowledge graphs.}
\label{tab:semantic_properties}
\end{table}

\begin{table}[H]
\centering
\small
\begin{tabular}{lcccccccccc}
\toprule
 & \rotatebox{90}{\textbf{Europeana}} & \rotatebox{90}{\textbf{Google Data Commons}} & \rotatebox{90}{\textbf{Bio2RDF}} & \rotatebox{90}{\textbf{British Museum ResearchSpace}} & \rotatebox{90}{\textbf{UniProt}} & \rotatebox{90}{\textbf{Wikidata}} & \rotatebox{90}{\textbf{EU ODP}} & \rotatebox{90}{\textbf{DBpedia}} & \rotatebox{90}{\textbf{LOV}} & \rotatebox{90}{\textbf{Nanopublications}} \\
\midrule
\textit{attribution} & $\times$ & $\times$ & $\times$ & $\times$ & $\times$ & $\times$ & $\times$ & $\times$ & $\times$ & $\times$ \\
\textit{source tracking} &  & $\times$ & $\times$ & $\times$ & $\times$ & $\times$ &  & $\times$ & $\times$ & $\times$ \\
\textit{data discovery} & $\times$ & $\times$ & $\times$ &  &  &  & $\times$ & $\times$ & $\times$ &  \\
\textit{knowledge curation} &  &  &  & $\times$ &  & $\times$ &  &  & $\times$ & $\times$ \\
\textit{reproducibility} &  &  & $\times$ &  & $\times$ &  & $\times$ &  &  & $\times$ \\
\textit{rights management} & $\times$ &  & $\times$ &  &  &  & $\times$ & $\times$ &  &  \\
\textit{cross-dataset integration} &  & $\times$ &  &  &  & $\times$ &  & $\times$ &  &  \\
\textit{data quality evaluation} &  &  &  &  & $\times$ & $\times$ &  &  &  & $\times$ \\
\textit{cross-portal interoperability} & $\times$ &  &  &  &  &  & $\times$ &  & $\times$ &  \\
\textit{scholarly citation} &  &  &  & $\times$ &  & $\times$ &  &  &  & $\times$ \\
\bottomrule
\end{tabular}
\caption{Semantic affordances of provenance across knowledge graphs.}
\label{tab:semantic_affordances}
\end{table}

\begin{table}[H]
\centering
\small
\begin{tabular}{lcccccccccc}
\toprule
 & \rotatebox{90}{\textbf{Europeana}} & \rotatebox{90}{\textbf{Google Data Commons}} & \rotatebox{90}{\textbf{Bio2RDF}} & \rotatebox{90}{\textbf{British Museum ResearchSpace}} & \rotatebox{90}{\textbf{UniProt}} & \rotatebox{90}{\textbf{Wikidata}} & \rotatebox{90}{\textbf{EU ODP}} & \rotatebox{90}{\textbf{DBpedia}} & \rotatebox{90}{\textbf{LOV}} & \rotatebox{90}{\textbf{Nanopublications}} \\
\midrule
\textit{PROV-O} &  &  &  &  &  &  & $\times$ & $\times$ & $\times$ & $\times$ \\
\textit{named graphs} &  &  & $\times$ &  &  &  &  & $\times$ &  & $\times$ \\
\textit{Wikidata model} &  &  &  &  &  & $\times$ &  &  &  &  \\
\textit{OAI-ORE aggregation} & $\times$ &  &  &  &  &  &  &  &  &  \\
\textit{Schema.org markup} &  & $\times$ &  &  &  &  &  &  &  &  \\
\textit{RDF-Star} &  &  &  &  & $\times$ &  &  &  &  &  \\
\textit{RDF reification} &  &  &  & $\times$ &  &  &  &  &  &  \\
\bottomrule
\end{tabular}
\caption{Pragmatic properties of provenance across knowledge graphs.}
\label{tab:pragmatic_properties}
\end{table}

\begin{table}[H]
\centering
\small
\begin{tabular}{lcccccccccc}
\toprule
 & \rotatebox{90}{\textbf{Europeana}} & \rotatebox{90}{\textbf{Google Data Commons}} & \rotatebox{90}{\textbf{Bio2RDF}} & \rotatebox{90}{\textbf{British Museum ResearchSpace}} & \rotatebox{90}{\textbf{UniProt}} & \rotatebox{90}{\textbf{Wikidata}} & \rotatebox{90}{\textbf{EU ODP}} & \rotatebox{90}{\textbf{DBpedia}} & \rotatebox{90}{\textbf{LOV}} & \rotatebox{90}{\textbf{Nanopublications}} \\
\midrule
\textit{SPARQL} & $\times$ & $\times$ & $\times$ & $\times$ & $\times$ & $\times$ & $\times$ & $\times$ & $\times$ & $\times$ \\
\textit{SPARQL endpoint} & $\times$ &  & $\times$ & $\times$ & $\times$ & $\times$ & $\times$ & $\times$ & $\times$ & $\times$ \\
\textit{bulk download} & $\times$ & $\times$ & $\times$ &  & $\times$ & $\times$ & $\times$ & $\times$ & $\times$ & $\times$ \\
\textit{linked data browsing} & $\times$ &  & $\times$ & $\times$ &  & $\times$ & $\times$ & $\times$ & $\times$ & $\times$ \\
\textit{REST API} & $\times$ & $\times$ &  & $\times$ & $\times$ & $\times$ & $\times$ & $\times$ & $\times$ &  \\
\textit{multilingual access} & $\times$ &  &  &  &  & $\times$ & $\times$ & $\times$ &  &  \\
\textit{SPARQL federation} &  &  &  &  & $\times$ & $\times$ &  & $\times$ &  &  \\
\textit{OWL DL reasoning} &  &  &  & $\times$ &  &  & $\times$ &  &  & $\times$ \\
\textit{SHACL} &  &  &  &  &  & $\times$ & $\times$ &  &  &  \\
\textit{natural language query} &  & $\times$ &  &  &  &  &  &  &  &  \\
\textit{SPARQL-Star} &  &  &  &  & $\times$ &  &  &  &  &  \\
\bottomrule
\end{tabular}
\caption{Pragmatic affordances of provenance across knowledge graphs.}
\label{tab:pragmatic_affordances}
\end{table}


\begin{figure}[H]
  \centering
  \adjustbox{max width=\textwidth, max totalheight=\textheight}{%
\begin{tikzpicture}[
    concept/.style={circle, draw, fill=white, inner sep=4pt, minimum size=16pt,
                    font=\small\bfseries},
    lbl/.style={font=\small}
]
  \draw (0.000,0.000) -- (-3.750,2.000);
  \draw (0.000,0.000) -- (-1.250,2.000);
  \draw (0.000,0.000) -- (1.250,2.000);
  \draw (0.000,0.000) -- (3.750,2.000);
  \draw (-3.750,2.000) -- (-3.750,12.000);
  \draw (-1.250,2.000) -- (-2.500,6.000);
  \draw (1.250,2.000) -- (-3.750,4.000);
  \draw (1.250,2.000) -- (-1.250,4.000);
  \draw (3.750,2.000) -- (1.250,4.000);
  \draw (3.750,2.000) -- (-3.750,4.000);
  \draw (3.750,2.000) -- (3.750,4.000);
  \draw (1.250,4.000) -- (0.000,6.000);
  \draw (1.250,4.000) -- (-2.500,10.000);
  \draw (-3.750,4.000) -- (2.500,6.000);
  \draw (-1.250,4.000) -- (-2.500,6.000);
  \draw (-1.250,4.000) -- (2.500,6.000);
  \draw (3.750,4.000) -- (0.000,6.000);
  \draw (3.750,4.000) -- (0.000,10.000);
  \draw (0.000,6.000) -- (-1.250,12.000);
  \draw (-2.500,6.000) -- (2.500,10.000);
  \draw (2.500,6.000) -- (0.000,8.000);
  \draw (0.000,8.000) -- (-2.500,10.000);
  \draw (0.000,8.000) -- (2.500,10.000);
  \draw (0.000,8.000) -- (0.000,10.000);
  \draw (-2.500,10.000) -- (1.250,12.000);
  \draw (-2.500,10.000) -- (-1.250,12.000);
  \draw (2.500,10.000) -- (-3.750,12.000);
  \draw (2.500,10.000) -- (1.250,12.000);
  \draw (2.500,10.000) -- (3.750,12.000);
  \draw (-3.750,12.000) -- (-2.500,14.000);
  \draw (-3.750,12.000) -- (0.000,14.000);
  \draw (0.000,10.000) -- (-1.250,12.000);
  \draw (0.000,10.000) -- (3.750,12.000);
  \draw (1.250,12.000) -- (-2.500,14.000);
  \draw (1.250,12.000) -- (2.500,14.000);
  \draw (-2.500,14.000) -- (0.000,16.000);
  \draw (-1.250,12.000) -- (2.500,14.000);
  \draw (3.750,12.000) -- (0.000,14.000);
  \draw (3.750,12.000) -- (2.500,14.000);
  \draw (0.000,14.000) -- (0.000,16.000);
  \draw (2.500,14.000) -- (0.000,16.000);
  \node[concept] (n0) at (0.000,0.000) {(a1)};
  \node[concept] (n1) at (-3.750,2.000) {(b1)};
  \node[concept] (n2) at (-1.250,2.000) {(c1)};
  \node[concept] (n3) at (1.250,2.000) {(d1)};
  \node[concept] (n4) at (3.750,2.000) {(e1)};
  \node[concept] (n5) at (1.250,4.000) {(h1)};
  \node[concept] (n6) at (-3.750,4.000) {(f1)};
  \node[concept] (n7) at (-1.250,4.000) {(g1)};
  \node[concept] (n8) at (3.750,4.000) {(i1)};
  \node[concept] (n9) at (0.000,6.000) {(k1)};
  \node[concept] (n10) at (-2.500,6.000) {(j1)};
  \node[concept] (n11) at (2.500,6.000) {(l1)};
  \node[concept] (n12) at (0.000,8.000) {(m1)};
  \node[concept] (n13) at (-2.500,10.000) {(n1)};
  \node[concept] (n14) at (2.500,10.000) {(p1)};
  \node[concept] (n15) at (-3.750,12.000) {(q1)};
  \node[concept] (n16) at (0.000,10.000) {(o1)};
  \node[concept] (n17) at (1.250,12.000) {(s1)};
  \node[concept] (n18) at (-2.500,14.000) {(u1)};
  \node[concept] (n19) at (-1.250,12.000) {(r1)};
  \node[concept] (n20) at (3.750,12.000) {(t1)};
  \node[concept] (n21) at (0.000,14.000) {(v1)};
  \node[concept] (n22) at (2.500,14.000) {(w1)};
  \node[concept] (n23) at (0.000,16.000) {(x1)};
\end{tikzpicture}
}%

  \caption{FCA lattice: pragmatic affordances. See Table~\ref{tab:fca-pragmatic_affordances} for concept legend.}
  \label{fig:fca-pragmatic_affordances}
\end{figure}
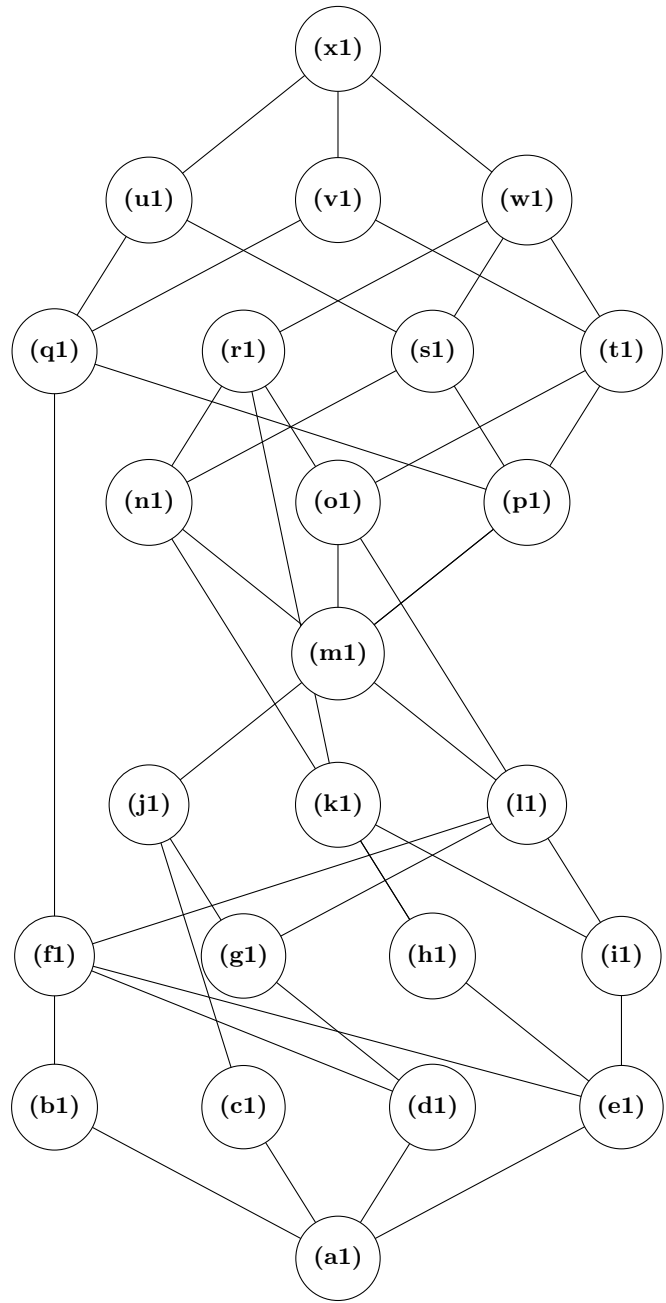

\small
\begin{longtable}{l p{6cm} p{6cm}}
  \caption{Concept legend for the FCA lattice in Figure~\ref{fig:fca-pragmatic_affordances}: pragmatic affordances} \\
  \label{tab:fca-pragmatic_affordances} \\
  \hline
  \textbf{ID} & \textbf{Objects} & \textbf{Attributes} \\
  \hline
  \endfirsthead
  \multicolumn{3}{l}{\small\itshape Concept legend for the FCA lattice in Figure~\ref{fig:fca-pragmatic_affordances}: pragmatic affordances (continued)} \\
  \hline
  \textbf{ID} & \textbf{Objects} & \textbf{Attributes} \\
  \hline
  \endhead
  \hline
  \multicolumn{3}{r}{\small\itshape Continued on next page} \\
  \endfoot
  \hline
  \endlastfoot
  (a1) & --- & SPARQL, SPARQL endpoint, REST API, linked data browsing, bulk download, multilingual access, natural language query, OWL DL reasoning, SPARQL-Star, SPARQL federation, SHACL \\
  (b1) & Google Data Commons & SPARQL, REST API, bulk download, natural language query \\
  (c1) & UniProt & SPARQL, SPARQL endpoint, REST API, bulk download, SPARQL-Star, SPARQL federation \\
  (d1) & Wikidata & SPARQL, SPARQL endpoint, REST API, linked data browsing, bulk download, multilingual access, SPARQL federation, SHACL \\
  (e1) & EU ODP & SPARQL, SPARQL endpoint, REST API, linked data browsing, bulk download, multilingual access, OWL DL reasoning, SHACL \\
  (f1) & Wikidata, EU ODP & SPARQL, SPARQL endpoint, REST API, linked data browsing, bulk download, multilingual access, SHACL \\
  (g1) & Wikidata, DBpedia & SPARQL, SPARQL endpoint, REST API, linked data browsing, bulk download, multilingual access, SPARQL federation \\
  (h1) & British Museum ResearchSpace, EU ODP & SPARQL, SPARQL endpoint, REST API, linked data browsing, OWL DL reasoning \\
  (i1) & EU ODP, Nanopublications & SPARQL, SPARQL endpoint, linked data browsing, bulk download, OWL DL reasoning \\
  (j1) & UniProt, Wikidata, DBpedia & SPARQL, SPARQL endpoint, REST API, bulk download, SPARQL federation \\
  (k1) & British Museum ResearchSpace, EU ODP, Nanopublications & SPARQL, SPARQL endpoint, linked data browsing, OWL DL reasoning \\
  (l1) & Europeana, Wikidata, EU ODP, DBpedia & SPARQL, SPARQL endpoint, REST API, linked data browsing, bulk download, multilingual access \\
  (m1) & Europeana, Wikidata, EU ODP, DBpedia, LOV & SPARQL, SPARQL endpoint, REST API, linked data browsing, bulk download \\
  (n1) & Europeana, British Museum ResearchSpace, Wikidata, EU ODP, DBpedia, LOV & SPARQL, SPARQL endpoint, REST API, linked data browsing \\
  (o1) & Europeana, Bio2RDF, Wikidata, EU ODP, DBpedia, LOV, Nanopublications & SPARQL, SPARQL endpoint, linked data browsing, bulk download \\
  (p1) & Europeana, UniProt, Wikidata, EU ODP, DBpedia, LOV & SPARQL, SPARQL endpoint, REST API, bulk download \\
  (q1) & Europeana, Google Data Commons, UniProt, Wikidata, EU ODP, DBpedia, LOV & SPARQL, REST API, bulk download \\
  (r1) & Europeana, Bio2RDF, British Museum ResearchSpace, Wikidata, EU ODP, DBpedia, LOV, Nanopublications & SPARQL, SPARQL endpoint, linked data browsing \\
  (s1) & Europeana, British Museum ResearchSpace, UniProt, Wikidata, EU ODP, DBpedia, LOV & SPARQL, SPARQL endpoint, REST API \\
  (t1) & Europeana, Bio2RDF, UniProt, Wikidata, EU ODP, DBpedia, LOV, Nanopublications & SPARQL, SPARQL endpoint, bulk download \\
  (u1) & Europeana, Google Data Commons, British Museum ResearchSpace, UniProt, Wikidata, EU ODP, DBpedia, LOV & SPARQL, REST API \\
  (v1) & Europeana, Google Data Commons, Bio2RDF, UniProt, Wikidata, EU ODP, DBpedia, LOV, Nanopublications & SPARQL, bulk download \\
  (w1) & Europeana, Bio2RDF, British Museum ResearchSpace, UniProt, Wikidata, EU ODP, DBpedia, LOV, Nanopublications & SPARQL, SPARQL endpoint \\
  (x1) & Europeana, Google Data Commons, Bio2RDF, British Museum ResearchSpace, UniProt, Wikidata, EU ODP, DBpedia, LOV, Nanopublications & SPARQL \\
\end{longtable}
\normalsize

\begin{figure}[H]
  \centering
  \adjustbox{max width=\textwidth, max totalheight=\textheight}{%
\begin{tikzpicture}[
    concept/.style={circle, draw, fill=white, inner sep=4pt, minimum size=16pt,
                    font=\small\bfseries},
    lbl/.style={font=\small}
]
  \draw (0.000,0.000) -- (-6.250,2.000);
  \draw (0.000,0.000) -- (-3.750,2.000);
  \draw (0.000,0.000) -- (-1.250,2.000);
  \draw (0.000,0.000) -- (1.250,2.000);
  \draw (0.000,0.000) -- (3.750,2.000);
  \draw (0.000,0.000) -- (6.250,2.000);
  \draw (-6.250,2.000) -- (0.000,6.000);
  \draw (-3.750,2.000) -- (0.000,6.000);
  \draw (-1.250,2.000) -- (0.000,6.000);
  \draw (1.250,2.000) -- (0.000,6.000);
  \draw (3.750,2.000) -- (0.000,6.000);
  \draw (6.250,2.000) -- (-1.250,4.000);
  \draw (6.250,2.000) -- (1.250,4.000);
  \draw (-1.250,4.000) -- (0.000,6.000);
  \draw (1.250,4.000) -- (0.000,6.000);
  \node[concept] (n0) at (0.000,0.000) {(a2)};
  \node[concept] (n1) at (-6.250,2.000) {(b2)};
  \node[concept] (n2) at (-3.750,2.000) {(c2)};
  \node[concept] (n3) at (-1.250,2.000) {(d2)};
  \node[concept] (n4) at (1.250,2.000) {(e2)};
  \node[concept] (n5) at (3.750,2.000) {(f2)};
  \node[concept] (n6) at (6.250,2.000) {(g2)};
  \node[concept] (n7) at (-1.250,4.000) {(h2)};
  \node[concept] (n8) at (1.250,4.000) {(i2)};
  \node[concept] (n9) at (0.000,6.000) {(j2)};
\end{tikzpicture}
}%

  \caption{FCA lattice: pragmatic properties. See Table~\ref{tab:fca-pragmatic_properties} for concept legend.}
  \label{fig:fca-pragmatic_properties}
\end{figure}
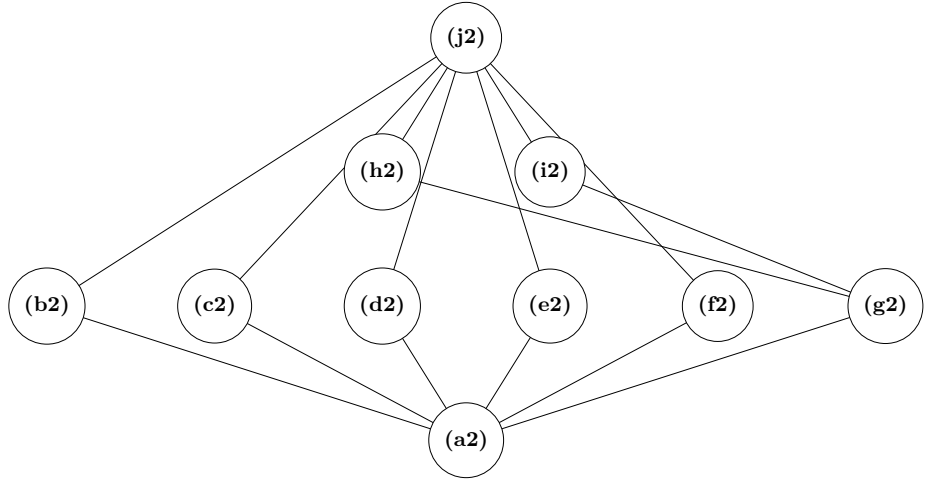

\small
\begin{longtable}{l p{6cm} p{6cm}}
  \caption{Concept legend for the FCA lattice in Figure~\ref{fig:fca-pragmatic_properties}: pragmatic properties} \\
  \label{tab:fca-pragmatic_properties} \\
  \hline
  \textbf{ID} & \textbf{Objects} & \textbf{Attributes} \\
  \hline
  \endfirsthead
  \multicolumn{3}{l}{\small\itshape Concept legend for the FCA lattice in Figure~\ref{fig:fca-pragmatic_properties}: pragmatic properties (continued)} \\
  \hline
  \textbf{ID} & \textbf{Objects} & \textbf{Attributes} \\
  \hline
  \endhead
  \hline
  \multicolumn{3}{r}{\small\itshape Continued on next page} \\
  \endfoot
  \hline
  \endlastfoot
  (a2) & --- & OAI-ORE aggregation, Schema.org markup, named graphs, RDF reification, RDF-Star, Wikidata model, PROV-O \\
  (b2) & Europeana & OAI-ORE aggregation \\
  (c2) & Google Data Commons & Schema.org markup \\
  (d2) & British Museum ResearchSpace & RDF reification \\
  (e2) & UniProt & RDF-Star \\
  (f2) & Wikidata & Wikidata model \\
  (g2) & DBpedia, Nanopublications & named graphs, PROV-O \\
  (h2) & Bio2RDF, DBpedia, Nanopublications & named graphs \\
  (i2) & EU ODP, DBpedia, LOV, Nanopublications & PROV-O \\
  (j2) & Europeana, Google Data Commons, Bio2RDF, British Museum ResearchSpace, UniProt, Wikidata, EU ODP, DBpedia, LOV, Nanopublications & --- \\
\end{longtable}
\normalsize

\begin{figure}[H]
  \centering
  \adjustbox{max width=\textwidth, max totalheight=\textheight}{%
\begin{tikzpicture}[
    concept/.style={circle, draw, fill=white, inner sep=4pt, minimum size=16pt,
                    font=\small\bfseries},
    lbl/.style={font=\small}
]
  \draw (0.000,0.000) -- (-6.250,2.000);
  \draw (0.000,0.000) -- (-3.750,2.000);
  \draw (0.000,0.000) -- (-1.250,2.000);
  \draw (0.000,0.000) -- (1.250,2.000);
  \draw (0.000,0.000) -- (3.750,2.000);
  \draw (0.000,0.000) -- (6.250,2.000);
  \draw (-6.250,2.000) -- (-6.250,4.000);
  \draw (-6.250,2.000) -- (-3.750,4.000);
  \draw (-6.250,2.000) -- (-7.500,6.000);
  \draw (-3.750,2.000) -- (-1.250,4.000);
  \draw (-3.750,2.000) -- (-5.000,6.000);
  \draw (-1.250,2.000) -- (1.250,4.000);
  \draw (-1.250,2.000) -- (-6.250,4.000);
  \draw (1.250,2.000) -- (3.750,4.000);
  \draw (1.250,2.000) -- (-3.750,4.000);
  \draw (3.750,2.000) -- (-2.500,6.000);
  \draw (3.750,2.000) -- (0.000,6.000);
  \draw (3.750,2.000) -- (-2.500,8.000);
  \draw (6.250,2.000) -- (6.250,4.000);
  \draw (6.250,2.000) -- (-1.250,4.000);
  \draw (1.250,4.000) -- (-2.500,6.000);
  \draw (1.250,4.000) -- (2.500,6.000);
  \draw (3.750,4.000) -- (-5.000,6.000);
  \draw (3.750,4.000) -- (0.000,6.000);
  \draw (-6.250,4.000) -- (2.500,6.000);
  \draw (-6.250,4.000) -- (0.000,8.000);
  \draw (-3.750,4.000) -- (2.500,6.000);
  \draw (-3.750,4.000) -- (0.000,6.000);
  \draw (6.250,4.000) -- (-7.500,6.000);
  \draw (6.250,4.000) -- (5.000,6.000);
  \draw (-1.250,4.000) -- (7.500,6.000);
  \draw (-1.250,4.000) -- (5.000,6.000);
  \draw (-2.500,6.000) -- (2.500,8.000);
  \draw (-5.000,6.000) -- (0.000,10.000);
  \draw (-7.500,6.000) -- (0.000,8.000);
  \draw (-7.500,6.000) -- (0.000,10.000);
  \draw (7.500,6.000) -- (-2.500,8.000);
  \draw (5.000,6.000) -- (0.000,10.000);
  \draw (2.500,6.000) -- (2.500,8.000);
  \draw (0.000,6.000) -- (2.500,8.000);
  \draw (0.000,6.000) -- (0.000,10.000);
  \draw (0.000,8.000) -- (0.000,12.000);
  \draw (-2.500,8.000) -- (0.000,10.000);
  \draw (2.500,8.000) -- (0.000,12.000);
  \draw (0.000,10.000) -- (0.000,12.000);
  \node[concept] (n0) at (0.000,0.000) {(a3)};
  \node[concept] (n1) at (-6.250,2.000) {(b3)};
  \node[concept] (n2) at (-3.750,2.000) {(c3)};
  \node[concept] (n3) at (-1.250,2.000) {(d3)};
  \node[concept] (n4) at (1.250,2.000) {(e3)};
  \node[concept] (n5) at (3.750,2.000) {(f3)};
  \node[concept] (n6) at (6.250,2.000) {(g3)};
  \node[concept] (n7) at (1.250,4.000) {(k3)};
  \node[concept] (n8) at (3.750,4.000) {(l3)};
  \node[concept] (n9) at (-6.250,4.000) {(h3)};
  \node[concept] (n10) at (-3.750,4.000) {(i3)};
  \node[concept] (n11) at (6.250,4.000) {(m3)};
  \node[concept] (n12) at (-1.250,4.000) {(j3)};
  \node[concept] (n13) at (-2.500,6.000) {(p3)};
  \node[concept] (n14) at (-5.000,6.000) {(o3)};
  \node[concept] (n15) at (-7.500,6.000) {(n3)};
  \node[concept] (n16) at (7.500,6.000) {(t3)};
  \node[concept] (n17) at (5.000,6.000) {(s3)};
  \node[concept] (n18) at (2.500,6.000) {(r3)};
  \node[concept] (n19) at (0.000,6.000) {(q3)};
  \node[concept] (n20) at (0.000,8.000) {(v3)};
  \node[concept] (n21) at (-2.500,8.000) {(u3)};
  \node[concept] (n22) at (2.500,8.000) {(w3)};
  \node[concept] (n23) at (0.000,10.000) {(x3)};
  \node[concept] (n24) at (0.000,12.000) {(y3)};
\end{tikzpicture}
}%

  \caption{FCA lattice: semantic affordances. See Table~\ref{tab:fca-semantic_affordances} for concept legend.}
  \label{fig:fca-semantic_affordances}
\end{figure}
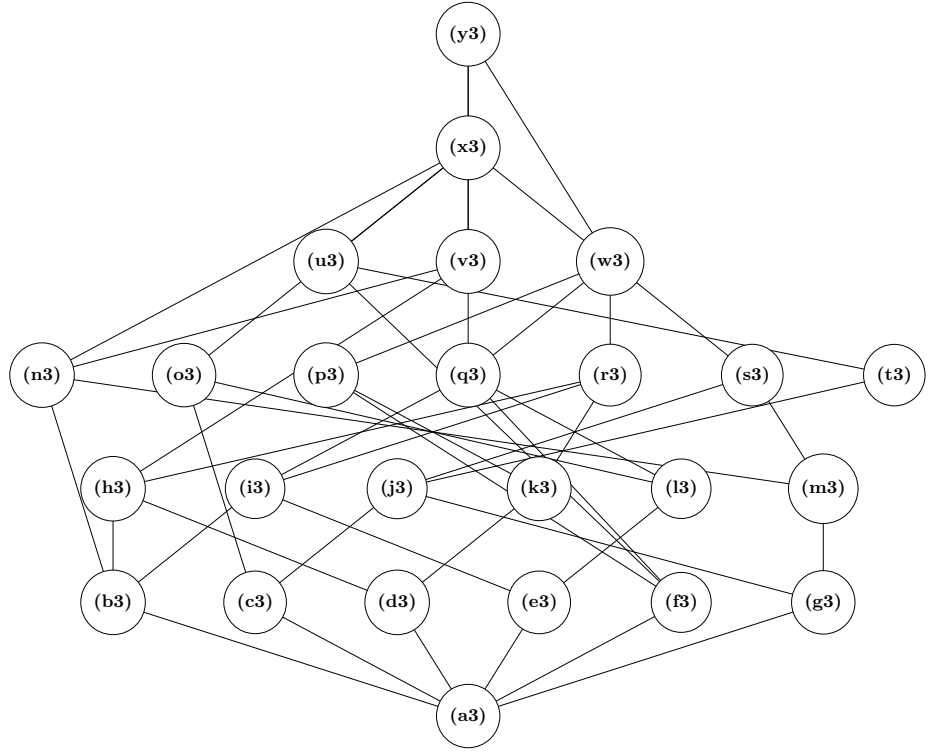

\small
\begin{longtable}{l p{6cm} p{6cm}}
  \caption{Concept legend for the FCA lattice in Figure~\ref{fig:fca-semantic_affordances}: semantic affordances} \\
  \label{tab:fca-semantic_affordances} \\
  \hline
  \textbf{ID} & \textbf{Objects} & \textbf{Attributes} \\
  \hline
  \endfirsthead
  \multicolumn{3}{l}{\small\itshape Concept legend for the FCA lattice in Figure~\ref{fig:fca-semantic_affordances}: semantic affordances (continued)} \\
  \hline
  \textbf{ID} & \textbf{Objects} & \textbf{Attributes} \\
  \hline
  \endhead
  \hline
  \multicolumn{3}{r}{\small\itshape Continued on next page} \\
  \endfoot
  \hline
  \endlastfoot
  (a3) & --- & attribution, data discovery, rights management, cross-portal interoperability, source tracking, cross-dataset integration, reproducibility, scholarly citation, knowledge curation, data quality evaluation \\
  (b3) & Bio2RDF & attribution, data discovery, rights management, source tracking, reproducibility \\
  (c3) & Wikidata & attribution, source tracking, cross-dataset integration, scholarly citation, knowledge curation, data quality evaluation \\
  (d3) & EU ODP & attribution, data discovery, rights management, cross-portal interoperability, reproducibility \\
  (e3) & DBpedia & attribution, data discovery, rights management, source tracking, cross-dataset integration \\
  (f3) & LOV & attribution, data discovery, cross-portal interoperability, source tracking, knowledge curation \\
  (g3) & Nanopublications & attribution, source tracking, reproducibility, scholarly citation, knowledge curation, data quality evaluation \\
  (h3) & Bio2RDF, EU ODP & attribution, data discovery, rights management, reproducibility \\
  (i3) & Bio2RDF, DBpedia & attribution, data discovery, rights management, source tracking \\
  (j3) & Wikidata, Nanopublications & attribution, source tracking, scholarly citation, knowledge curation, data quality evaluation \\
  (k3) & Europeana, EU ODP & attribution, data discovery, rights management, cross-portal interoperability \\
  (l3) & Google Data Commons, DBpedia & attribution, data discovery, source tracking, cross-dataset integration \\
  (m3) & UniProt, Nanopublications & attribution, source tracking, reproducibility, data quality evaluation \\
  (n3) & Bio2RDF, UniProt, Nanopublications & attribution, source tracking, reproducibility \\
  (o3) & Google Data Commons, Wikidata, DBpedia & attribution, source tracking, cross-dataset integration \\
  (p3) & Europeana, EU ODP, LOV & attribution, data discovery, cross-portal interoperability \\
  (q3) & Google Data Commons, Bio2RDF, DBpedia, LOV & attribution, data discovery, source tracking \\
  (r3) & Europeana, Bio2RDF, EU ODP, DBpedia & attribution, data discovery, rights management \\
  (s3) & UniProt, Wikidata, Nanopublications & attribution, source tracking, data quality evaluation \\
  (t3) & British Museum ResearchSpace, Wikidata, Nanopublications & attribution, source tracking, scholarly citation, knowledge curation \\
  (u3) & British Museum ResearchSpace, Wikidata, LOV, Nanopublications & attribution, source tracking, knowledge curation \\
  (v3) & Bio2RDF, UniProt, EU ODP, Nanopublications & attribution, reproducibility \\
  (w3) & Europeana, Google Data Commons, Bio2RDF, EU ODP, DBpedia, LOV & attribution, data discovery \\
  (x3) & Google Data Commons, Bio2RDF, British Museum ResearchSpace, UniProt, Wikidata, DBpedia, LOV, Nanopublications & attribution, source tracking \\
  (y3) & Europeana, Google Data Commons, Bio2RDF, British Museum ResearchSpace, UniProt, Wikidata, EU ODP, DBpedia, LOV, Nanopublications & attribution \\
\end{longtable}
\normalsize

\begin{figure}[H]
  \centering
  \adjustbox{max width=\textwidth, max totalheight=\textheight}{%
\begin{tikzpicture}[
    concept/.style={circle, draw, fill=white, inner sep=4pt, minimum size=16pt,
                    font=\small\bfseries},
    lbl/.style={font=\small}
]
  \draw (0.000,0.000) -- (-7.500,2.000);
  \draw (0.000,0.000) -- (-5.000,2.000);
  \draw (0.000,0.000) -- (-2.500,2.000);
  \draw (0.000,0.000) -- (0.000,2.000);
  \draw (0.000,0.000) -- (2.500,2.000);
  \draw (0.000,0.000) -- (5.000,2.000);
  \draw (0.000,0.000) -- (7.500,2.000);
  \draw (-7.500,2.000) -- (-7.500,4.000);
  \draw (-5.000,2.000) -- (-5.000,4.000);
  \draw (-5.000,2.000) -- (-3.750,8.000);
  \draw (-2.500,2.000) -- (-2.500,4.000);
  \draw (-2.500,2.000) -- (0.000,4.000);
  \draw (0.000,2.000) -- (-5.000,4.000);
  \draw (0.000,2.000) -- (2.500,4.000);
  \draw (0.000,2.000) -- (-7.500,6.000);
  \draw (2.500,2.000) -- (-2.500,4.000);
  \draw (5.000,2.000) -- (-7.500,4.000);
  \draw (5.000,2.000) -- (0.000,4.000);
  \draw (5.000,2.000) -- (5.000,4.000);
  \draw (7.500,2.000) -- (7.500,4.000);
  \draw (7.500,2.000) -- (2.500,4.000);
  \draw (7.500,2.000) -- (-5.000,6.000);
  \draw (-7.500,4.000) -- (-2.500,6.000);
  \draw (-7.500,4.000) -- (0.000,6.000);
  \draw (-5.000,4.000) -- (2.500,6.000);
  \draw (-5.000,4.000) -- (-2.500,10.000);
  \draw (-2.500,4.000) -- (0.000,6.000);
  \draw (0.000,4.000) -- (5.000,6.000);
  \draw (0.000,4.000) -- (0.000,6.000);
  \draw (7.500,4.000) -- (7.500,6.000);
  \draw (5.000,4.000) -- (-2.500,6.000);
  \draw (5.000,4.000) -- (5.000,6.000);
  \draw (5.000,4.000) -- (-7.500,6.000);
  \draw (5.000,4.000) -- (-5.000,6.000);
  \draw (-2.500,6.000) -- (-1.250,8.000);
  \draw (-2.500,6.000) -- (-3.750,8.000);
  \draw (5.000,6.000) -- (1.250,8.000);
  \draw (5.000,6.000) -- (-3.750,8.000);
  \draw (2.500,4.000) -- (2.500,6.000);
  \draw (2.500,4.000) -- (7.500,6.000);
  \draw (-7.500,6.000) -- (-1.250,8.000);
  \draw (-7.500,6.000) -- (1.250,8.000);
  \draw (-7.500,6.000) -- (3.750,8.000);
  \draw (-5.000,6.000) -- (3.750,8.000);
  \draw (0.000,6.000) -- (-3.750,8.000);
  \draw (-1.250,8.000) -- (-2.500,10.000);
  \draw (-1.250,8.000) -- (0.000,10.000);
  \draw (2.500,6.000) -- (0.000,12.000);
  \draw (1.250,8.000) -- (-2.500,10.000);
  \draw (1.250,8.000) -- (2.500,10.000);
  \draw (7.500,6.000) -- (3.750,8.000);
  \draw (-3.750,8.000) -- (-2.500,10.000);
  \draw (3.750,8.000) -- (0.000,10.000);
  \draw (3.750,8.000) -- (2.500,10.000);
  \draw (-2.500,10.000) -- (0.000,12.000);
  \draw (0.000,10.000) -- (0.000,12.000);
  \draw (2.500,10.000) -- (0.000,12.000);
  \node[concept] (n0) at (0.000,0.000) {(a4)};
  \node[concept] (n1) at (-7.500,2.000) {(b4)};
  \node[concept] (n2) at (-5.000,2.000) {(c4)};
  \node[concept] (n3) at (-2.500,2.000) {(d4)};
  \node[concept] (n4) at (0.000,2.000) {(e4)};
  \node[concept] (n5) at (2.500,2.000) {(f4)};
  \node[concept] (n6) at (5.000,2.000) {(g4)};
  \node[concept] (n7) at (7.500,2.000) {(h4)};
  \node[concept] (n8) at (-7.500,4.000) {(i4)};
  \node[concept] (n9) at (-5.000,4.000) {(j4)};
  \node[concept] (n10) at (-2.500,4.000) {(k4)};
  \node[concept] (n11) at (0.000,4.000) {(l4)};
  \node[concept] (n12) at (7.500,4.000) {(o4)};
  \node[concept] (n13) at (5.000,4.000) {(n4)};
  \node[concept] (n14) at (-2.500,6.000) {(r4)};
  \node[concept] (n15) at (5.000,6.000) {(u4)};
  \node[concept] (n16) at (2.500,4.000) {(m4)};
  \node[concept] (n17) at (-7.500,6.000) {(p4)};
  \node[concept] (n18) at (-5.000,6.000) {(q4)};
  \node[concept] (n19) at (0.000,6.000) {(s4)};
  \node[concept] (n20) at (-1.250,8.000) {(x4)};
  \node[concept] (n21) at (2.500,6.000) {(t4)};
  \node[concept] (n22) at (1.250,8.000) {(y4)};
  \node[concept] (n23) at (7.500,6.000) {(v4)};
  \node[concept] (n24) at (-3.750,8.000) {(w4)};
  \node[concept] (n25) at (3.750,8.000) {(z4)};
  \node[concept] (n26) at (-2.500,10.000) {(aa4)};
  \node[concept] (n27) at (0.000,10.000) {(ab4)};
  \node[concept] (n28) at (2.500,10.000) {(ac4)};
  \node[concept] (n29) at (0.000,12.000) {(ad4)};
\end{tikzpicture}
}%

  \caption{FCA lattice: semantic properties. See Table~\ref{tab:fca-semantic_properties} for concept legend.}
  \label{fig:fca-semantic_properties}
\end{figure}
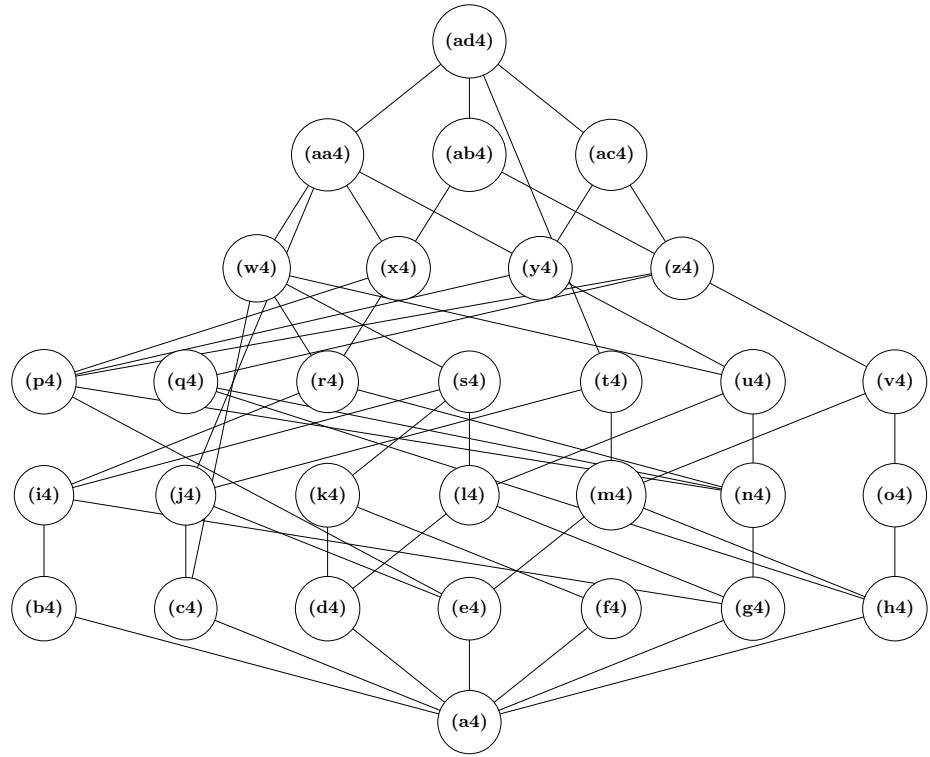

\small
\begin{longtable}{l p{6cm} p{6cm}}
  \caption{Concept legend for the FCA lattice in Figure~\ref{fig:fca-semantic_properties}: semantic properties} \\
  \label{tab:fca-semantic_properties} \\
  \hline
  \textbf{ID} & \textbf{Objects} & \textbf{Attributes} \\
  \hline
  \endfirsthead
  \multicolumn{3}{l}{\small\itshape Concept legend for the FCA lattice in Figure~\ref{fig:fca-semantic_properties}: semantic properties (continued)} \\
  \hline
  \textbf{ID} & \textbf{Objects} & \textbf{Attributes} \\
  \hline
  \endhead
  \hline
  \multicolumn{3}{r}{\small\itshape Continued on next page} \\
  \endfoot
  \hline
  \endlastfoot
  (a4) & --- & authorship, data provider, aggregation chain, licensing, dataset-level lineage, source organisation, triple-level lineage, conversion process, temporal information, scholarly assertion, epistemic status, evidence type, agent roles, multilinguality \\
  (aa4) & Europeana, Google Data Commons, Bio2RDF, Wikidata, EU ODP, DBpedia, LOV & data provider \\
  (ab4) & Europeana, British Museum ResearchSpace, UniProt, Wikidata, EU ODP, LOV, Nanopublications & authorship \\
  (ac4) & Bio2RDF, British Museum ResearchSpace, UniProt, Wikidata, EU ODP, LOV, Nanopublications & temporal information \\
  (ad4) & Europeana, Google Data Commons, Bio2RDF, British Museum ResearchSpace, UniProt, Wikidata, EU ODP, DBpedia, LOV, Nanopublications & --- \\
  (b4) & Europeana & authorship, data provider, aggregation chain, licensing, dataset-level lineage \\
  (c4) & Google Data Commons & data provider, dataset-level lineage, source organisation, triple-level lineage \\
  (d4) & Bio2RDF & data provider, licensing, dataset-level lineage, conversion process, temporal information \\
  (e4) & Wikidata & authorship, data provider, triple-level lineage, temporal information, epistemic status, evidence type \\
  (f4) & DBpedia & data provider, licensing, dataset-level lineage, conversion process, multilinguality \\
  (g4) & LOV & authorship, data provider, licensing, dataset-level lineage, temporal information, agent roles \\
  (h4) & Nanopublications & authorship, triple-level lineage, temporal information, scholarly assertion, epistemic status, evidence type, agent roles \\
  (i4) & Europeana, LOV & authorship, data provider, licensing, dataset-level lineage \\
  (j4) & Google Data Commons, Wikidata & data provider, triple-level lineage \\
  (k4) & Bio2RDF, DBpedia & data provider, licensing, dataset-level lineage, conversion process \\
  (l4) & Bio2RDF, LOV & data provider, licensing, dataset-level lineage, temporal information \\
  (m4) & UniProt, Wikidata, Nanopublications & authorship, triple-level lineage, temporal information, epistemic status, evidence type \\
  (n4) & EU ODP, LOV & authorship, data provider, dataset-level lineage, temporal information, agent roles \\
  (o4) & British Museum ResearchSpace, Nanopublications & authorship, temporal information, scholarly assertion, epistemic status \\
  (p4) & Wikidata, EU ODP, LOV & authorship, data provider, temporal information \\
  (q4) & EU ODP, LOV, Nanopublications & authorship, temporal information, agent roles \\
  (r4) & Europeana, EU ODP, LOV & authorship, data provider, dataset-level lineage \\
  (s4) & Europeana, Bio2RDF, DBpedia, LOV & data provider, licensing, dataset-level lineage \\
  (t4) & Google Data Commons, UniProt, Wikidata, Nanopublications & triple-level lineage \\
  (u4) & Bio2RDF, EU ODP, LOV & data provider, dataset-level lineage, temporal information \\
  (v4) & British Museum ResearchSpace, UniProt, Wikidata, Nanopublications & authorship, temporal information, epistemic status \\
  (w4) & Europeana, Google Data Commons, Bio2RDF, EU ODP, DBpedia, LOV & data provider, dataset-level lineage \\
  (x4) & Europeana, Wikidata, EU ODP, LOV & authorship, data provider \\
  (y4) & Bio2RDF, Wikidata, EU ODP, LOV & data provider, temporal information \\
  (z4) & British Museum ResearchSpace, UniProt, Wikidata, EU ODP, LOV, Nanopublications & authorship, temporal information \\
\end{longtable}
\normalsize

\end{onlyextended}
\end{document}